%% file: colm2026_conference.tex
\documentclass{article} 
\usepackage[final]{colm2026_conference}

\include{packages}

\usepackage{microtype}
\usepackage{hyperref}
\usepackage{url}
\usepackage{booktabs}

\usepackage{lineno}

\definecolor{darkblue}{rgb}{0, 0, 0.5}
\hypersetup{colorlinks=true, citecolor=darkblue, linkcolor=darkblue, urlcolor=darkblue}

\title{Agentic Test-Time Scaling for WebAgents}

\author{Nicholas Lee\thanks{Equal contribution. $^{\dagger}$Corresponding author: \texttt{amirgh@berkeley.edu}}$\ \ ^{1}$,
  Lutfi Eren Erdogan$^{*1}$,
  Chris Joseph John$^{1}$,
  Surya Krishnapillai$^{1}$, \\
  \textbf{Michael W. Mahoney$^{1,2,3}$,
  Kurt Keutzer$^{1}$,
  Amir Gholami$^{\dagger 1,2}$}
  \\
  $^{1}$UC Berkeley\quad
  $^{2}$ICSI\quad
  $^{3}$LBNL
}

\begin{document}

\ifcolmsubmission
\linenumbers
\fi

\maketitle

\input{text/_0_abstract}

\input{text/_1_intro}
\input{text/_2_related_work}
\input{text/_3_study}
\input{text/_4_discussion}

\input{text/_5_conclusion}

\section*{Acknowledgements}
We acknowledge the gracious support from the Furiosa AI, Intel, Apple, NVIDIA, Macronix, and Mozilla team.
Furthermore, we appreciate the support from
Google Cloud, the Google TRC team Prof.~David Patterson, along with support from Google Gemini team, and Divy Thakkar.
Prof.~Keutzer's lab is also sponsored by funding through BDD and BAIR.
We also acknowledge support by the Director, Office of Science, Office of Advanced Scientific Computing Research, of the U.S. Department of Energy under Contract No. DE-AC02-05CH11231.
MWM would also like to acknowledge DARPA, DOE, NSF, and ONR.
DOE SciGPT grant. Our conclusions do not necessarily reflect the position or the policy of our sponsors, and no official endorsement should be~inferred.

\bibliography{colm2026_conference}
\bibliographystyle{colm2026_conference}

\newpage

\appendix
\input{text/_6_appendix}

\end{document}

%% file: packages.tex
\usepackage{wrapfig}

\usepackage{mathpazo}
\usepackage[utf8]{inputenc}
\usepackage{verbatim}
\usepackage{calc}
\usepackage{mathrsfs}
\usepackage{mathtools}
\usepackage{url}
\usepackage{algorithm2e}
\usepackage{amsmath}
\usepackage{amssymb}
\usepackage{graphicx}
\usepackage{hyperref}
\usepackage{xcolor}
\usepackage{comment}
\usepackage{soul}
\usepackage{booktabs}

\definecolor{forestgreen}{rgb}{0.13, 0.55, 0.13}

\usepackage{tabularx}
\usepackage{multirow}
\usepackage{multicol}
\usepackage{subcaption}

\usepackage{colortbl}
\usepackage{tabulary}
\usepackage{etoolbox}

\definecolor{colorA}{RGB}{189,201,225}
\definecolor{colorB}{RGB}{103,169,207}
\definecolor{colorC}{RGB}{ 28,144,153}
\definecolor{colorD}{RGB}{  1,108, 89}

\newcolumntype{R}{>{\columncolor{gray!40}}r}
\newcolumntype{L}{>{\columncolor{gray!40}}l}
\newcolumntype{C}{>{\columncolor{gray!40}}c}

\newcommand{\OURS}{\textsc{CATTS}\xspace}

\usepackage{microtype}
\usepackage{graphicx}
\usepackage{booktabs} 

\usepackage{enumitem}

\newcommand\appref{Appendix~\ref}
\newcommand\eref{Eq.~\ref}
\newcommand\fref{Figure~\ref}
\newcommand\tref{Table~\ref}
\newcommand\sref{Section~\ref}

\usepackage{amssymb}
\usepackage{pifont}
\definecolor{darkerred}{RGB}{190, 0, 0} 

\newcommand{\BASEMODEL}{\texttt{gpt-oss-120b}}
\newcommand{\JUDGEMODEL}{\texttt{Qwen3-VL-30B-A3B-Instruct}}

%% file: text/_0_abstract.tex
\begin{abstract}
Test-time scaling has become a standard way to improve performance and boost reliability of neural network models.
However, its behavior on \emph{agentic}, multi-step tasks remains less well-understood: small per-step errors can compound over long horizons; and we find that naive policies that uniformly increase sampling show diminishing returns.
In this work, we present \OURS, a simple technique for dynamically allocating compute for multi-step agents.
We first conduct an empirical study of inference-time scaling for web agents.
We find that uniformly increasing per-step compute quickly saturates in long-horizon environments.
We then investigate stronger aggregation strategies, including an LLM-based Arbiter that can outperform naive voting, but that can overrule high-consensus decisions.
We show that uncertainty statistics derived from the agent's own vote distribution (entropy and top-1/top-2 margin) correlate with downstream success and provide a practical signal for dynamic compute allocation.
Based on these findings, we introduce Confidence-Aware Test-Time Scaling (\OURS), which uses vote-derived uncertainty to allocate compute only when decisions are genuinely contentious. \OURS improves performance on WebArena-Lite, Online-Mind2Web, and GoBrowse by up to 11.8\% over majority voting while using fewer tokens than uniform scaling, providing both efficiency gains and an interpretable decision rule.

\end{abstract}

%% file: text/_1_intro.tex
\section{Introduction}
Large language models (LLMs) are increasingly used not only to produce text, but to \emph{take actions}: they can call tools, navigate websites, operate software, and execute multi-step procedures in interactive environments.
In these settings, an agent must repeatedly choose a next action, e.g., clicking a button, typing into a form, or issuing a search query, based on the current observation and the history of what it has already done \citep{patil2024gorilla, schick2023toolformer, erdogan2024tinyagent}.
The sequential nature of these tasks makes reliability much more challenging than simpler single-shot question answering: a single poor decision can send the trajectory into an unrecoverable state, and small per-step error rates can compound over many steps \citep{erdogan2025plan}.

A widely used strategy for improving LLM capabilities at test time is \emph{test-time scaling} \citep{snell2024scaling, brown2024large, cobbe2021training, zelikman2022star}. The core idea is to spend compute generating more tokens at test-time, rather than spending this compute during pretraining. For single-shot reasoning tasks, this can yield large gains because different samples can explore different reasoning paths: even if the first attempt fails, additional attempts often contain a correct solution, and majority voting \citep{wang2022self} or verification \citep{shinn2023reflexion} can exploit this diversity.

An important question that remains to be answered is: \emph{What does inference-time scaling look like for multi-step, tool-using agents?}
A direct analogue of the single-shot recipe is to consider each step separately, treating each step the same way one would treat an ordinary reasoning task.
At each step $t$, instead of generating one action, we can sample $N$ candidate actions from the base model and then choose which action to execute.
Repeating this procedure throughout a trajectory yields a simple and appealing ``knob'' for scaling up compute \citep{wang2022self, rsa2025}.

However, a naive strategy, \emph{uniform scaling}, where we always sample the same number of candidates and apply the same selection rule at every step, runs into two issues:
(1)~\textbf{Wasted computation on easy steps}: many steps are obvious (form fills, clicking submit), making additional sampling redundant;
(2)~\textbf{High-variance decisions}: when votes spread across many distinct options, majority voting provides weak signal.

A common remedy is to add an LLM-based \emph{arbiter} that reasons over the candidate set \citep{rsa2025, zhang2024generative, muennighoff2025s1}.
However, arbiters are prone to overthinking \citep{cuadron2025danger}: they can override correct high-consensus decisions, making extra compute harmful rather than helpful.
This raises a key question: \emph{where} should we spend additional inference-time compute?

We study this design space systematically by adapting several test-time scaling techniques (voting, arbitration, and confidence-aware filtering \citep{fu2026deep}) to long-horizon web agents.
The outcome is a simple principle: \emph{inference-time compute should be allocated where it is likely to change the decision}.
We find that the distribution of answers generated at each step predicts when additional selection helps (\sref{sec:uncertainty_diagnostic}).

Using this insight, we present \OURS (\textsc{Confidence-Aware Test-Time Scaling}), which uses vote-derived uncertainty to dynamically allocate compute only when decisions are genuinely contentious.
\OURS achieves consistent improvements on WebArena-Lite and Online-Mind2Web (up to $+11.8\%$ over majority voting) while using fewer tokens than uniform scaling.

\begin{figure}[t]
\centering
\vspace{-0.1in}
\includegraphics[width=0.9\textwidth, trim=10 175 10 100, clip]{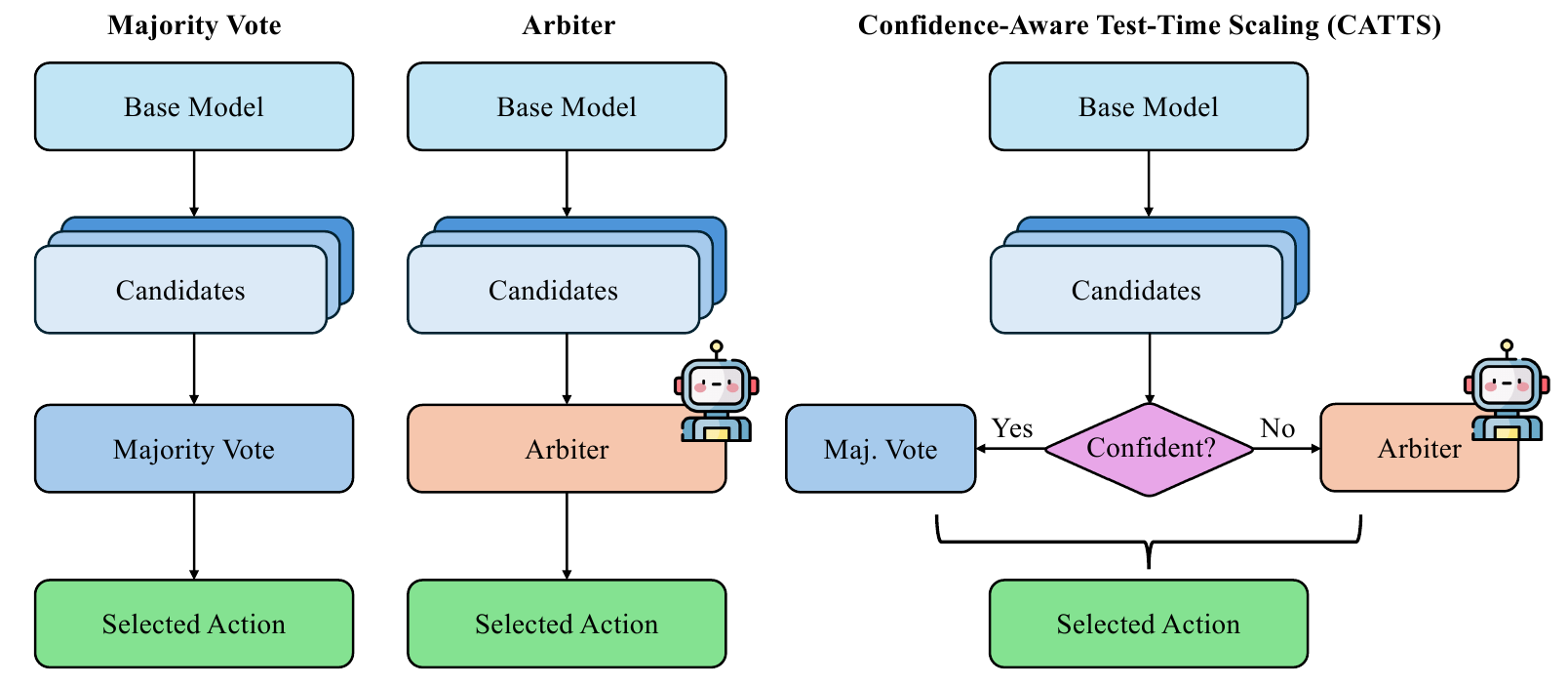}
\caption{\textbf{Comparing agentic inference-time scaling methods.} Visual comparison of selection strategies at each step. \textbf{Left:} Majority Voting samples $N$ candidates and selects the most frequent action via argmax over vote distribution $p_t(a)$. \textbf{Center-Left:} Arbiter samples $N$ candidates and uses an additional LLM call to reason over candidates and select the best action. \textbf{Center-Right:} \OURS conditionally invokes the arbiter when vote-derived uncertainty (entropy $H_t$ or margin $\Delta_t$) exceeds threshold $\tau$.
}
\vspace{-0.1in}
\label{fig:method_comparison}
\end{figure}

%% file: text/_2_related_work.tex
\section{Related Work}
\label{sec:related_work}

\paragraph{Inference-Time Scaling and Test-Time Compute.}
Vast amounts of recent work have explored how to improve the performance and reliability of LLMs by spending additional test-time compute.
Self-consistency decoding \citep{wang2022self} demonstrated substantial gains on reasoning tasks by sampling multiple chain-of-thought traces \citep{wei2022chain, kojima2022large} and taking a majority vote over final answers, treating it as an ensemble problem. Subsequent work has explored richer aggregation strategies including ranked voting and diversity-aware selection \citep{wang2025ranked, naik2023diversity, resource_rational2024}, although several studies highlight fundamental limits when sampled outputs share correlated errors \citep{byerly2026self, turpin2023language}.

Alternative approaches allocate compute via structured search over reasoning steps \citep{yao2023tree, besta2024graph, lats2024_zhou, kim2024llm} or use confidence-aware filtering to gate computation \citep{fu2026deep, know_what_they_know2022_kadavath}. Recent work formalizes compute-optimal policies that trade off parallel sampling and sequential refinement \citep{snell2024scaling}, connecting to adaptive computation ideas in neural models \citep{act2016}. Another related line of work is DeepConf \citep{fu2026deep}, which uses the model's own confidence scores to dynamically filter out uncertain and low-quality chain-of-thought. 
The token-level probability distribution is used to compute confidence scores which can be used for confidence-weighted majority voting or filtering.

Our work applies these principles to sequential decision-making in long-horizon tool-use settings. We use vote-derived uncertainty to gate arbitration at each step, enabling dynamic allocation without requiring token-level confidence access. Unlike prior confidence-filtering approaches that operate on single-shot reasoning traces, we derive uncertainty from agreement among multiple sampled candidates, and we operate online in environments where errors compound across steps.

\vspace{-0.05in}
\paragraph{Tool-Using Agents and Long-Horizon Tasks.}
Recent progress in building LLM-driven agents has been driven by frameworks that interleave reasoning and tool execution. ReAct \citep{yao2023react} demonstrated that intermixing reasoning with tool invocations improves performance, while subsequent work introduced structured decomposition such as decoupling planning from execution \citep{erdogan2025plan, rewoo2023_zhou, erdogan2024tinyagent} or leveraging code for numerical reasoning \citep{pal2022_gao, program_of_thoughts2022_chen}. To enable robust tool use, several frameworks teach models to invoke external APIs through self-supervised learning \citep{schick2023toolformer}, retrieval-aware training \citep{patil2024gorilla, toolllm2023_qin}, or structured integration layers \citep{mrkl2022_karpas, li2023api}. Complementary lines of work improve agent policies through reinforcement learning \citep{workflow_guided_exploration2018_liu}, iterative refinement via self-reflection \citep{shinn2023reflexion, zelikman2022star}, or retrieval-augmented reasoning \citep{rarr2023_gao}.

Evaluation of these methods relies on increasingly realistic benchmarks. For web navigation, environments range from early workflow-based tasks \citep{worldofbits2017_shi} to modern suites spanning e-commerce, forums, and content management \citep{zhou2024webarena, visualwebarena2024_koh, deng2023mind2web, yao2022webshop, webgpt2021_nakano, workarena2024, workarena_pp2024}, while computer-use benchmarks extend to desktops and OS-level interaction \citep{xie2024osworld, rawles2023androidinthewild, rawles2024androidworld, appagent2024_zhang, tinyclick2024_pawlowski}. Multi-domain suites evaluate generalization across diverse tasks \citep{agentbench2023_liu, stabletoolbench2024_guo, voyager2023_wang, generative_agents2023_park, bagel2024_yang}.

Multi-agent debate and voting strategies \citep{multiagent_debate2023_du} and extreme decomposition via massive redundancy \citep{mdap2025} offer alternative paths to reliability, often at prohibitive computational costs. 
Our work is orthogonal to these training-based and architectural improvements: we assume a fixed base agent, and we focus on how to allocate inference-time compute efficiently at test time. 
The central motivation is practical deployment, where unnecessary model calls translate directly to latency, cost, and energy use, making adaptive allocation essential for real-world viability.

%% file: text/_3_study.tex
\section{From Static to Dynamic Inference-Time Scaling}
\label{sec:narrative_core}

This section presents the main study on inference-time scaling techniques for agents.
We start from the most direct adaptation of inference-time scaling to web agents: sampling multiple candidate actions and aggregating them, progressively refining the selection mechanism based on empirical failure modes.
We find that uniformly allocating inference-time compute inconsistently and inefficiently affects performance, and that the distribution of the actions themselves is an effective test-time signal for performance.
This insight is used to propose a simple and interpretable inference-time scaling policy that dynamically allocates compute only when the candidate action distribution is uncertain.

\subsection{Experimental Setup}
\label{sec:experimental_setup}
We study tool-using web agents that interact with a browser over multiple steps. The underlying agent is prompted with ReAct \citep{yao2023react} and uses \BASEMODEL{} as the base model.
At each step $t$, the agent observes a cleaned HTML representation of the current page state and outputs a structured action space of 8 tools including \textbf{click}, \textbf{type\_text}, \textbf{scroll}, etc.
The full action space, agent prompts, conversation format, and error handling mechanisms are detailed in Appendices \ref{app:design_decisions} and \ref{app:agent_prompts}.

We evaluate on WebArena-Lite \citep{liu2025visualagentbench}, which has 165 tasks with programmatic success checks, and Online-Mind2Web \citep{xue2025an}, which has 300 tasks across diverse real-world websites evaluated using an LLM-as-a-judge \citep{zheng2023judging} protocol with \JUDGEMODEL{}.
Online-Mind2Web features substantially longer trajectories (mean 17.4 steps) than WebArena-Lite (8--12 steps), providing a challenging testbed for long-horizon scaling.
We also evaluate on GoBrowse \citep{gobrowse2025} (341 tasks); full results are provided in \appref{app:gobrowse_results}. A detailed comparison of benchmark characteristics and LLM-as-judge reliability is provided in \appref{app:benchmark_characteristics}.
We measure compute cost using the total token count per task, defined as the sum of prompt (input) and completion (output) tokens across all model calls, as a vast majority of the tokens consumed come from the input prompt.
All results on WebArena-Lite are averaged across three different seeds unless otherwise noted; Online-Mind2Web results are single-run due to compute constraints.

\paragraph{Action Clustering and Vote Distributions.}

At each time step $t$, the agent observes $o_t$ and samples $N$ candidate actions from the base model:
\begin{equation}
\tilde{a}^{(i)}_t \sim M(\,\cdot \mid o_t\,), \quad i=1,\dots,N.
\end{equation}
We parse candidates into structured actions to obtain a set of action clusters $\mathcal{A}_t$. Let $n_t(a)$ denote the number of candidates assigned to cluster $a\in\mathcal{A}_t$.
This induces a distribution over the sampled actions
\begin{equation}
p_t(a) = \frac{n_t(a)}{N},
\end{equation}
which parameterizes how much the sampled candidates agree on each distinct action at step $t$.

One problem that arises when trying to do majority voting in this action space is that different outputs are semantically the same, but have different text representations (e.g., ``\texttt{N/A}'' vs.\ ``\texttt{Not found}'').
We therefore use a lightweight semantic deduplicator LLM to cluster candidate actions into semantically equivalent buckets before counting votes (See \appref{app:agent_prompts} for the prompt and \appref{app:deduplication_ablation} for additional analysis).

The central question of our study is this: \emph{Given the current observation $o_t$ and a multi-set of candidates $\{\tilde{a}^{(i)}_t\}_{i=1}^N$, how should we select the executed action $a_t$ to maximize task success under a token budget?}
We begin with static aggregation rules (majority voting and uniform arbitration) and then show how vote-derived uncertainty enables a dynamic allocation of computation. Figure~\ref{fig:method_comparison} provides a visual overview of the selection strategies we evaluate.

\subsection{Static Baseline: Majority Voting}
\label{sec:static_voting}

The most direct adaptation of inference-time scaling to an agent loop is to sample multiple candidate next actions and execute the action with the most support. Concretely, after clustering candidates into $\mathcal{A}_t$, the executed action is
\begin{equation}
a_t \;=\; \arg\max_{a \in \mathcal{A}_t} \; p_t(a).
\label{eq:majority_vote}
\end{equation}
We refer to this as \emph{majority voting} (or frequency voting).
Most importantly, the compute allocation is uniform: the same number of candidates $N$ is sampled at every step, and the same selection rule in \eref{eq:majority_vote} is applied throughout the~trajectory.

\paragraph{Observation: Majority Vote Yields Diminishing Returns.}
As we can see in \tref{tab:static_scaling}, simply increasing candidate-generation compute yields diminishing or even negative returns.
On WebArena-Lite, scaling from $N{=}1$ to $N{=}10$ improves success from 38.8\% to 43.2\%, but doubling from $N{=}10$ to $N{=}20$ actually decreases success by 0.2\% despite doubling tokens.
On Online-Mind2Web, majority voting actually \emph{degrades} with more candidates (22.6\% at $N{=}1$ to 15.1\% at $N{=}20$), likely due to vote splitting across semantically similar but textually distinct actions in longer trajectories.
We also evaluated Plan-and-Act style agents \citep{erdogan2025plan} with factorized candidate-generation budgets across plan and action sampling, and the results exhibit similar non-monotonic scaling behavior, establishing that the issue affects different agent architectures (see \appref{app:plan_and_act} for full results).

\begin{table}[t]
  \centering
  \caption{\textbf{Static inference-time scaling with majority vote.} On WebArena-Lite, the final doubling ($N{=}10 \to 20$) produces negligible gain. On Online-Mind2Web, majority voting \emph{degrades} with more candidates (22.6\% $\to$ 15.1\%).}
  \label{tab:static_scaling}
  \vspace{1mm}
  \small
  \begin{tabular}{@{}lcccc@{}}
    \toprule
    & \multicolumn{2}{c}{WebArena-Lite} & \multicolumn{2}{c}{Online-Mind2Web} \\
    \cmidrule(lr){2-3} \cmidrule(lr){4-5}
    Budget & Success & Tokens & Success & Tokens \\
    \midrule
    $N{=}1$ & 38.8\% & 96K & 22.6\% & 169K \\
    $N{=}5$ & 42.4\% & 460K & 20.2\% & 888K \\
    \rowcolor{orange!12}
    $N{=}10$ & \textbf{43.2\%} & 920K & 17.9\% & 1.7M \\
    $N{=}20$ & 43.0\% & 1.8M & 15.1\% & 3.6M \\
    \bottomrule
  \end{tabular}
\end{table}

\paragraph{Takeaway.}
For long-horizon tool use, uniformly scaling up compute is often inefficient. Empirically, we observe that some steps produce highly diverse candidate sets where votes are spread across many distinct actions rather than converging on a clear winner. In these uncertain regimes, simple vote counting provides weak signal for selecting the correct action. This motivates selection mechanisms that can reason about ambiguous candidate sets, which we explore next.

\subsection{From Voting to Arbitration}
\label{sec:arbiter}

When majority vote produces diverse, uncertain candidate sets, a natural idea is to introduce an additional inference-time mechanism that \emph{reasons over the candidate set}.
Concretely, after sampling $N$ candidates and clustering into $\mathcal{A}_t$, we construct a short prompt containing the current observation $o_t$ and a list of representative actions (one per cluster).
An \emph{arbiter} model then chooses the action to execute:
\begin{equation}
a_t \;=\; \textsc{Arbiter}(o_t, \mathcal{A}_t, \{n_t(a)\}_{a\in\mathcal{A}_t}).
\label{eq:arbiter_select}
\end{equation}
In all experiments, the arbiter uses the same model as the agent (\BASEMODEL{}), and semantic deduplication is applied prior to selection. See \appref{app:agent_prompts} for implementation~details.

\paragraph{Observation: Arbitration Improves Over Majority Vote.}
Replacing frequency voting with an arbiter yields a consistent improvement over static majority voting across both benchmarks in our setting.
Intuitively, the arbiter can use the observation context to break ties among plausible actions and to discard candidates that look superficially popular but are contextually inappropriate.
\tref{tab:arbiter_scaling} summarizes the comparison between (i) majority vote, (ii) a single arbiter call per step, and (iii) scaled arbitration described below.

\paragraph{Arbiter Scaling.}
Because the arbiter model itself is an LLM call, we can also apply inference-time scaling \emph{to the selection mechanism}. With ``Arbiter Scaling,'' we query $K$ independent arbiters and aggregate their chosen actions by majority~vote:
\begin{equation}
a_t = \arg\max_{a \in \mathcal{A}_t} \frac{1}{K}\sum_{j=1}^K \mathbb{I}\!\left[\textsc{Select}_j(\cdot){=}a\right].
\label{eq:arbiter_scaling}
\end{equation}
Empirically, increasing $K$ yields further gains beyond a single arbiter call, and it consistently outperforms spending an equivalent token budget solely on increasing $N$ under majority vote.

\begin{table}[t]
  \centering
  \caption{\textbf{Arbiter scaling ($N{=}5$).} On WebArena-Lite, scaling improves to 44.6\% ($K{=}10$) but degrades at $K{=}20$. On Online-Mind2Web, arbitration provides large gains ($+7.2\%$ at $K{=}1$), with further improvements from scaling.}
  \label{tab:arbiter_scaling}
  \vspace{1mm}
  \small
  \begin{tabular}{@{}lcccc@{}}
    \toprule
    & \multicolumn{2}{c}{WebArena-Lite} & \multicolumn{2}{c}{Online-Mind2Web} \\
    \cmidrule(lr){2-3} \cmidrule(lr){4-5}
    Method & Success & Tokens & Success & Tokens \\
    \midrule
    Majority vote ($N{=}5$) & 42.4\% & 460K & 20.2\% & 888K \\
    Arbiter ($K{=}1$) & 42.8\% & 442K & 27.4\% & 960K \\
    Arbiter scaling ($K{=}5$) & 44.2\% & 645K & 26.3\% & 1.4M \\
    Arbiter scaling ($K{=}10$) & \textbf{44.6\%} & 899K & \textbf{30.7\%} & 2.1M \\
    Arbiter scaling ($K{=}20$) & 42.0\% & 1.4M & 29.3\% & 2.8M \\
    \bottomrule
  \end{tabular}
\end{table}

\paragraph{Observation: Arbitration is not Uniformly Beneficial.}
Although arbitration improves average performance, we also observe regimes where an additional arbiter call reduces success, particularly when the candidate set already exhibits strong agreement.
For example, in \tref{tab:arbiter_scaling}, increasing from $K{=}10$ to $K{=}20$ at $N{=}5$ causes performance to \emph{drop} from 44.6\% to 42.0\%. \sref{sec:uncertainty_diagnostic} analyzes this failure mode in detail and shows it is predictable from vote-derived uncertainty statistics.
See \appref{app:arbiter_scaling_full} for full results.

\paragraph{Deeper Aggregation Methods.}
We also evaluated Recursive Self-Aggregation (RSA) \citep{rsa2025}, which iteratively refines candidates over multiple rounds.
Despite substantially higher compute (up to 80 LLM calls per step vs.\ 10--11 for majority vote and single arbiter), RSA achieved comparable but not better performance than simpler methods.
Thus, we focus on single-round arbitration for the remainder of our study; full RSA results are provided in \appref{app:rsa_full}.

\subsection{DeepConf-Style Confidence Filtering}
\label{sec:deepconf}
We also evaluated DeepConf \citep{fu2026deep}, which uses token-level confidence scores for filtering and weighted voting.
The best DeepConf variant (average trace, $N{=}10$) achieves 43.8\% on WebArena-Lite, a modest gain over majority voting (43.2\%).
On Online-Mind2Web, DeepConf provides limited gains (18.3--24.7\%), substantially underperforming arbitration (27.4\% at $K{=}1$).
Critically, DeepConf requires access to token-level log probabilities, limiting applicability to API-only models.
Full results are in \appref{app:deepconf_full}.
In the next section, we explore an alternative signal derived purely from the vote distribution that achieves stronger benefits while remaining applicable to any model.

%% file: text/_4_discussion.tex
\section{Using Vote-Derived Uncertainty as a Test-Time Signal}
\label{sec:uncertainty_diagnostic}

We saw (in \sref{sec:arbiter}) that replacing frequency voting with an arbiter improves average performance, and that scaling the arbiters can improve performance further (see \tref{tab:arbiter_scaling}).
However, closer inspection of trajectories revealed a consistent pattern: the arbiter is most helpful on \emph{contentious} steps, where multiple plausible actions compete, but it can be harmful on \emph{high-consensus} steps where candidates already agree. 
On such steps, arbitration risks overriding a correct consensus action (see \fref{fig:arbiter_failure}).
This motivates the main question of this subsection: \emph{Can we distinguish between regimes where arbitration improves decisions and where it introduces harmful overrides?}

\subsection{Analyzing Action Distributions}
Given the vote distribution $p_t(\cdot)$ over action clusters $\mathcal{A}_t$, we compute two uncertainty statistics:
\begin{align}
H_t \;&=\; - \sum_{a \in \mathcal{A}_t} p_t(a)\,\log p_t(a), \label{eq:entropy}\\
\Delta_t \;&=\; p_t(a^{(1)}_t) - p_t(a^{(2)}_t), \label{eq:margin}
\end{align}
where $a^{(1)}_t$ and $a^{(2)}_t$ are the highest- and second-highest-probability clusters under $p_t(\cdot)$. Intuitively, $H_t$ measures overall disagreement (higher entropy means votes are spread across multiple options), while $\Delta_t$ measures decisiveness (larger gaps indicate a clear winner). High-consensus regimes correspond to \emph{low} $H_t$ and \emph{high} $\Delta_t$, contentious regimes correspond to \emph{high} $H_t$ and \emph{low} $\Delta_t$.

\begin{figure}[t]
  \centering
  \includegraphics[width=0.85\linewidth]{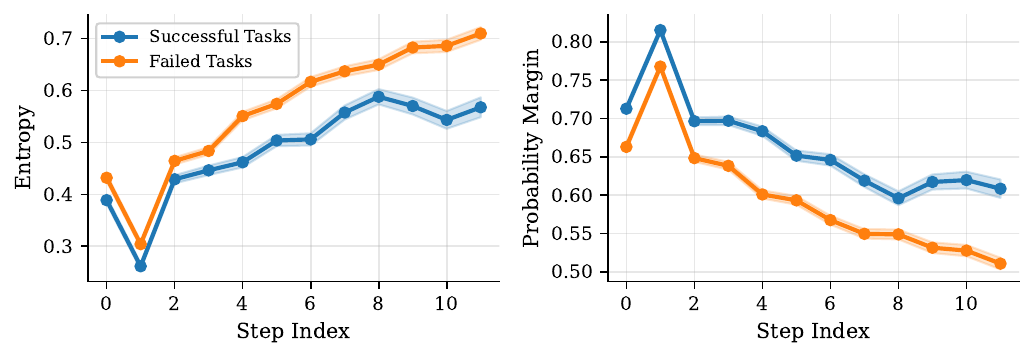}
  \caption{\textbf{Uncertainty profiles over trajectory steps.} Entropy $H_t$ (top) and probability margin $\Delta_t$ (bottom) versus step index, separated by successful (blue) and failed (orange) runs. Failed tasks consistently exhibit higher entropy and lower margins, with the gap widening at later steps.}
  \label{fig:uncertainty_over_steps}
\end{figure}

\paragraph{Uncertainty profiles differ between successful and failed trajectories.}
We compare successful and failed tasks by their uncertainty statistics across steps. Let $T$ be the number of steps in a trajectory, and define task-level averages
\begin{equation}
\bar{H} = \frac{1}{T}\sum_{t=1}^{T} H_t,
\qquad
\bar{\Delta} = \frac{1}{T}\sum_{t=1}^{T} \Delta_t.
\end{equation}
Empirically, successful trajectories tend to exhibit lower entropy and higher margins, whereas failed trajectories show the opposite trend. \fref{fig:uncertainty_over_steps} plots $H_t$ and $\Delta_t$ against the index of each step, separately for successful and failed runs, showing that disagreement typically spikes at pivotal decision points and that these spikes are more pronounced (or more frequent) in failed runs.

\paragraph{Uncertainty predicts when arbitration helps versus hurts.}
The failure case in \fref{fig:arbiter_failure} illustrates this distinction: at the pivotal step, the sampled candidates exhibit overwhelming agreement (vote split 9/10), corresponding to a large margin $\Delta_t$ and low entropy $H_t$. In this high-consensus regime, the arbiter overrides the dominant cluster and chooses a minority alternative that derails the trajectory. A frequency-based rule would instead execute the consensus action, which in our logs for this task, leads to success.

\begin{figure}[t]  
 \vspace{-0.3in}
  \centering
  \begin{subfigure}[t]{0.43\linewidth}
    \centering
    \includegraphics[width=\linewidth]{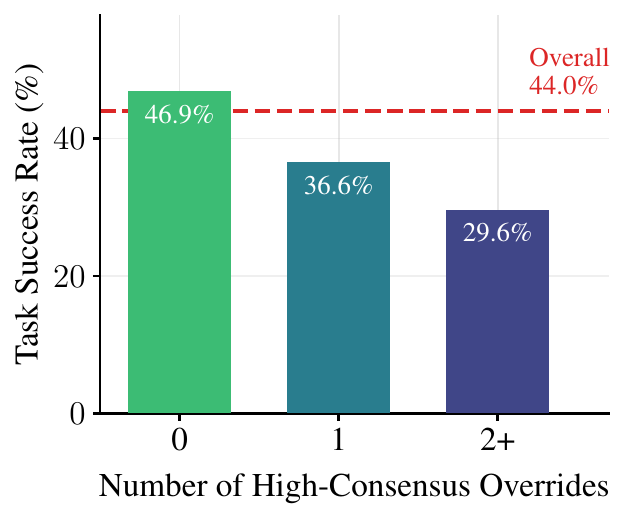}
    \phantomsubcaption\label{fig:override_analysis}
  \end{subfigure}
  \hfill
  \begin{subfigure}[t]{0.43\linewidth}
    \centering
    \includegraphics[width=\linewidth]{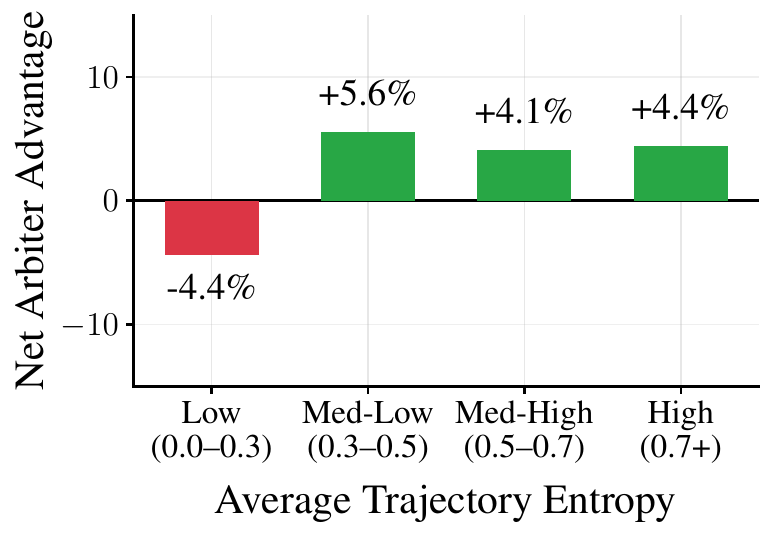}
    \phantomsubcaption\label{fig:arbiter_by_entropy}
  \end{subfigure}
  \vspace{-0.1in}
  \caption{\textbf{Arbiter effectiveness depends on vote-derived uncertainty.}
  \textbf{(Left)} Task success rate decreases as the number of high-consensus overrides ($\Delta_t > 0.7$) increases, showing a dose-response pattern ($p = 0.026$, Fisher's exact test). High-consensus overrides harm performance: tasks where the arbiter overrides majority vote on high-margin steps ($\Delta_t > 0.7$) show a significant dose-response decline in success rate.
  \textbf{(Right)} Net arbiter advantage by entropy bin shows that arbitration hurts at low entropy ($-4.4\%$) but helps at high entropy ($+4$-$6\%$). Grouping tasks by average trajectory entropy reveals that arbitration is harmful at low entropy but beneficial at high entropy, motivating uncertainty-gated compute allocation.}
  \label{fig:arbiter_analysis}
\end{figure}

To quantify this pattern, we analyzed all 495 task-runs from our arbiter experiments, classifying each task by whether the arbiter overrode the majority vote on any high-consensus step ($\Delta_t > 0.7$). \fref{fig:override_analysis} summarizes the results. Tasks without high-consensus overrides succeed at 46.9\%, compared to 35.0\% for tasks with at least one such override: a significant 11.9\% difference ($p = 0.026$, Fisher's exact test). This effect exhibits a dose-response pattern: tasks with zero overrides succeed at 46.9\%, those with exactly one at 36.6\%, and those with two or more at 29.6\%. The pattern is consistent across all five website categories.

Conversely, we find that arbitration provides benefit when uncertainty is high. \fref{fig:arbiter_by_entropy} groups tasks by average trajectory entropy and compares arbiter versus majority voting outcomes across all $N$ configurations. At low entropy (0.0-0.3), the arbiter shows a net disadvantage of $-4.4\%$, consistent with the override failures above. However, at higher entropy levels, the arbiter yields positive net advantages ($+4$-$6\%$), demonstrating that arbitration adds value when votes are diverse and majority voting lacks a clear~signal.

These findings illustrate a core principle: when $\Delta_t$ is high and $H_t$ is low, the model is already confident, and arbitration introduces override risk, rather than improving the decision; conversely, when entropy is high and votes are diverse, arbitration can reason over the candidate set and improve selection. 
This motivates a dynamic allocation rule: invoke arbitration primarily when vote-derived uncertainty indicates that the decision is genuinely contentious.

\subsection{\OURS: Confidence-Aware Test-Time Scaling}
\label{sec:adis}

We use vote-derived uncertainty to decide whether to spend extra compute on arbitration at each step. Let $U_t$ be a scalar uncertainty score computed from the vote distribution $p_t(\cdot)$. \OURS applies a threshold $\tau$ to gate arbitration:
\begin{equation}
a_t =
\begin{cases}
\arg\max_{a} p_t(a), & U_t \le \tau \\[4pt]
\textsc{Arbiter}(o_t, \mathcal{A}_t, \{n_t(a)\}_{a\in\mathcal{A}_t}), & U_t > \tau.
\end{cases}
\label{eq:adis_policy}
\end{equation}
We instantiate $U_t$ in two effective variants:
\begin{align}
U^{(\mathrm{ent})}_t &= H_t, \quad
U^{(\mathrm{mrg})}_t = 1 - \Delta_t.
\end{align}
Both entropy-gated and margin-gated \OURS improved the accuracy--compute tradeoff in our experiments. For both settings, we tuned $\tau$ via running a simple grid over thresholds (e.g., $\tau \in \{0.2,0.3,\dots,0.8\}$) and report results across this sweep (with the best-performing setting indicated where relevant).

\subsection{Results}
\label{sec:results}
\begin{figure}[t]
  \vspace{-0.3in}
  \centering
  \includegraphics[width=\textwidth]{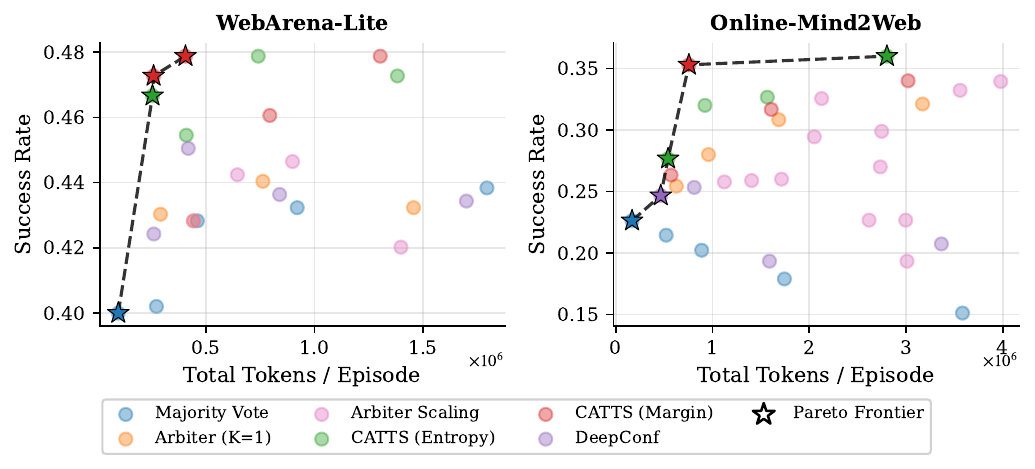}
  \caption{\textbf{Accuracy--compute frontier across all methods.}
  Success rate versus total tokens per episode on WebArena-Lite (left) and Online-Mind2Web (right). \OURS achieves Pareto improvements on both benchmarks: on WebArena-Lite, it reaches 47.9\% at ${\sim}400$K tokens; on Online-Mind2Web, it reaches 32.0\% at 922K tokens, outperforming always-arbitrate while using fewer tokens.}
  \label{fig:frontier}
\end{figure}

\OURS consistently improves over static baselines across both benchmarks. \fref{fig:frontier} plots success rate versus total tokens per episode for majority voting, uniform arbitration, DeepConf, and \OURS.
The key outcome is that \OURS achieves higher success than majority voting at comparable or lower token budgets, while using fewer tokens than always-arbitrate for higher success levels.
On WebArena-Lite, margin-gated \OURS achieves 47.9\% at only 405K tokens, a $+5.5\%$ gain over majority voting while using 12\% \emph{fewer} tokens.
On Online-Mind2Web, entropy-gated \OURS ($\tau{=}0.7$) achieves 32.0\%, a $+11.8\%$ gain over majority voting and $+4.6\%$ over always-arbitrate.
Averaging across all thresholds in our sweep on WebArena-Lite, \OURS achieves 45.6\%, which is still a consistent 2.4\% gain that does not depend on threshold selection (See \appref{app:threshold_sensitivity} for more analysis).
We additionally conducted a bootstrap calibration study (\appref{app:threshold_calibration}) showing that threshold selection requires minimal tuning: with only 10\% of tasks as calibration data, regret is ${\leq}1.1\%$ on both benchmarks.

Because \OURS invokes the arbiter only on uncertain steps, it reduces the number of arbitration calls substantially. In our experiments, the arbiter is invoked on approximately 40--60\% of steps on average, concentrating compute on the steps most likely to benefit.
This selective invocation also yields token savings: on WebArena-Lite, margin-gated \OURS uses 12\% fewer tokens than majority voting (405K vs.\ 460K) and 8\% fewer than always-arbitrate (442K), while on Online-Mind2Web it uses 4\% fewer tokens than always-arbitrate (922K vs.\ 960K).

\begin{table}[t]
\vspace{-0.05in}
  \centering
  \caption{\textbf{\OURS results ($N{=}5$).} On WebArena-Lite, margin-gated \OURS achieves 47.9\% at only 405K tokens, fewer than majority voting (460K). On Online-Mind2Web, entropy-gated \OURS achieves 32.0\% ($+11.8\%$ over majority vote, $+4.6\%$ over always-arbitrate).}
  \label{tab:adis_summary}
  \vspace{1mm}
  \small
  \begin{tabular}{@{}lcccc@{}}
    \toprule
    & \multicolumn{2}{c}{WebArena-Lite} & \multicolumn{2}{c}{Online-Mind2Web} \\
    \cmidrule(lr){2-3} \cmidrule(lr){4-5}
    Method & Success & Tokens & Success & Tokens \\
    \midrule
    Majority vote & 42.4\% & 460K & 20.2\% & 888K \\
    Always-arbitrate & 42.8\% & 442K & 27.4\% & 960K \\
    \rowcolor{green!12}
    \OURS ($H$, best $\tau$) & 45.5\% & 409K & \textbf{32.0\%} & 922K \\
    \rowcolor{green!12}
    \OURS ($\Delta$, best $\tau$) & \textbf{47.9\%} & 405K & 30.0\% & 906K \\
    \bottomrule
  \end{tabular}
  \vspace{-0.2in}
\end{table}

Our experiments reveal that per-step candidate sampling falls into two regimes.
\textbf{Regime 1: Redundancy.} Many steps are routine; candidates concentrate on a single cluster (low entropy, high margin). Additional sampling produces duplicates, and arbitration introduces override risk (\fref{fig:consensus_profile}, Appendices~\ref{app:vote_distribution} and~\ref{app:failure_examples}).
\textbf{Regime 2: Contention.} A smaller number of steps act as trajectory pivots with diffuse vote distributions. These are where selection quality matters most, as errors compound rather than self-correct. Arbitration adds value by reasoning over plausible candidates.
These gains generalize across model families, including Gemini 3.1 Flash Lite and Qwen3.5-122B-A10B (\appref{app:additional_models}).

%% file: text/_5_conclusion.tex
\section{Conclusion}
\label{sec:conclusion}

We present a systematic study of inference-time scaling for long-horizon web agents.
We find that naive per-step scaling shows diminishing or negative returns, and that LLM-based arbitration helps on contentious steps but can override correct high-consensus decisions.
Vote-derived uncertainty statistics (entropy and margin) reliably distinguish these regimes.
\OURS uses this signal to dynamically allocate compute, achieving consistent improvements across WebArena-Lite, Online-Mind2Web, and GoBrowse (up to $+11.8\%$ over majority voting) while using fewer tokens.

%% file: text/_6_appendix.tex
\section{Agent Design Decisions}
\label{app:design_decisions}

\subsection{Action Space}

Our agents use a minimal action space with 8 tools designed for web navigation:

\begin{itemize}[noitemsep]
\item \textbf{click}(element\_id): Click on a DOM element by its unique identifier
\item \textbf{type\_text}(element\_id, text): Enter text into an input field without submitting
\item \textbf{hover}(element\_id): Hover over an element to reveal tooltips or dropdown menus
\item \textbf{scroll}(direction): Scroll the page up or down to reveal more content
\item \textbf{select\_dropdown\_option}(element\_id, value): Select an option from a dropdown menu
\item \textbf{search}(element\_id, text): Type text and submit (equivalent to type + Enter)
\item \textbf{go\_back}(): Navigate to the previous page in browser history
\item \textbf{exit}(message): Terminate the episode with a final answer or status message
\end{itemize}

All element-targeting tools require an \texttt{element\_id} parameter that corresponds to an injected integer \texttt{id} attribute in the DOM. This simplifies element identification compared to CSS selectors or XPaths.

\subsection{Error Handling and Validation}

The agent framework includes several error checks that provide feedback during the retry loop:

\begin{itemize}[noitemsep]
\item \textbf{MustCallExactlyOneToolCheck}: Ensures exactly one function call per step
\item \textbf{InvalidActionSchemaCheck}: Validates that function arguments match the expected schema
\item \textbf{ElementMustExistCheck}: Verifies that referenced element IDs exist in the current DOM
\item \textbf{MustProvideReasoningCheck}: Requires the agent to provide reasoning in the analysis channel
\item \textbf{RepeatingActionLoopCheck}: Detects and prevents repeated ineffective actions
\end{itemize}

When a check fails, the agent receives an error message and has up to 5 retry attempts to produce a valid action.

\subsection{Conversation Format}

Each agent turn follows a structured format with the GPT-OSS Harmony interface:

\begin{enumerate}[noitemsep]
\item \textbf{System message}: Establishes agent role and channel usage
\item \textbf{Developer message}: Provides detailed instructions and constraints
\item \textbf{User message}: Contains the task, current HTML state, and any error feedback
\item \textbf{Assistant response}: Analysis in the \texttt{analysis} channel, function call in \texttt{commentary}
\item \textbf{Tool observation}: Result of the action with updated HTML state
\end{enumerate}

The conversation history preserves prior reasoning and actions but only includes full HTML in the most recent tool observation to manage context length. This allows the agent to reference its own reasoning trajectory while grounding decisions in the current page state.

\section{Agent Prompts}
\label{app:agent_prompts}

This section provides the key prompts used in our system components.

\subsection{Base Agent (ReAct Executor)}

The base agent uses a channel-based prompting format with separate channels for reasoning (\texttt{analysis}) and action emission (\texttt{commentary}). The system prompt establishes the agent's role:

\begin{quote}
\small
\texttt{You are a web-execution agent that completes browser tasks by emitting function calls in each step.}

\texttt{How each step works:}
\begin{itemize}[noitemsep,topsep=0pt]
\item \texttt{Inputs you'll see each step: User task (overall goal), prior reasoning and function calls, latest HTML snapshot of current page state.}
\item \texttt{What you must produce: First write \textbf{analysis} (detailed reasoning), then emit \textbf{exactly one function call} in \textbf{commentary}.}
\end{itemize}

\texttt{Channels: \textbf{analysis} $\rightarrow$ deep reasoning; \textbf{commentary} $\rightarrow$ one function call.}
\end{quote}

The developer prompt provides detailed instructions for element targeting, function-calling discipline, and recovery from errors:

\begin{quote}
\small
\texttt{Element targeting via numeric id (must-follow):}
\begin{itemize}[noitemsep,topsep=0pt]
\item \texttt{The environment injects a special \texttt{id} attribute that is an integer string (e.g., ``15'').}
\item \texttt{When a function requires an element, supply this integer id exactly as shown in the latest HTML.}
\item \texttt{Do not provide CSS selectors, XPaths, inner text, or made-up ids.}
\end{itemize}

\texttt{Recovery and loop-avoidance:}
\begin{itemize}[noitemsep,topsep=0pt]
\item \texttt{Do not repeat an ineffective action more than once.}
\item \texttt{After an error or no-op, re-analyze the latest HTML before choosing a different action.}
\end{itemize}
\end{quote}

\subsection{Semantic Deduplicator}

The semantic deduplicator clusters equivalent actions before voting. Its prompt instructs:

\begin{quote}
\small
\texttt{You are a deterministic semantic action deduplicator for a web-navigation system.}

\texttt{Context:}
\begin{itemize}[noitemsep,topsep=0pt]
\item \texttt{All inputs in one call are the same action type (TYPE | SEARCH | STOP).}
\item \texttt{All inputs act on the same element target (already checked upstream).}
\item \texttt{Your task: group semantically equivalent actions and pick a representative per group.}
\end{itemize}

\texttt{What ``equivalent'' means:}
\begin{itemize}[noitemsep,topsep=0pt]
\item \texttt{Same intent after trivial normalization (case, spacing, punctuation) and minor paraphrase/synonymy.}
\item \texttt{Do NOT merge items that change intent/topic (e.g., ``apple store'' $\neq$ ``apple id login'').}
\item \texttt{If unsure, keep them separate (prefer false negatives over false positives).}
\end{itemize}

\texttt{Output format: Clusters: [[rep\_index, other\_idx, ...], [rep\_index2, ...]]}
\end{quote}

\subsection{Arbiter}

The arbiter selects the best action from deduplicated candidates. Its prompt specifies:

\begin{quote}
\small
\texttt{You are an expert web-navigation action arbiter.}

\texttt{You will receive: User intent, previous actions, current page state (cleaned HTML), candidate actions.}

\texttt{Decision Criteria:}
\begin{enumerate}[noitemsep,topsep=0pt]
\item \texttt{Progress toward intent: Prefer actions that concretely advance the task}
\item \texttt{Relevance: Target visible, relevant elements from the current HTML}
\item \texttt{Avoid repetition: Don't repeat failed or redundant actions}
\item \texttt{Context awareness: Consider current page state and previous actions}
\item \texttt{Feasibility: Ensure the action is executable on the current page}
\end{enumerate}

\texttt{Output Format:}\\
\texttt{Thoughts: <reasoning about which action is best>}\\
\texttt{Pick: <number from 1 to N>}\\
\texttt{Confidence: <decimal from 0.0 to 1.0>}
\end{quote}

\subsection{External Planner (Plan-and-Act)}

For the Plan-and-Act baseline, we use a two-agent system where an external planner generates high-level plans and a separate executor agent follows them. The planner's system prompt establishes its role:

\begin{quote}
\small
\texttt{You are a web-navigation agent that normally completes browser tasks by thinking in an \textbf{analysis} channel and then emitting exactly one function call (using the provided tools) per step.}

\texttt{In this role, you are acting as the \textbf{Planner} in a two-agent system:}

\texttt{- You see the overall \textbf{User task}, your earlier \textbf{plans}, the executed \textbf{tool calls}, and the latest \textbf{HTML snapshot} (inside the most recent tool observation).}

\texttt{- A separate \textbf{Executor} agent will later read your plan and actually perform the function calls in the browser.}

\texttt{For this planner turn:}

\texttt{- Use the \textbf{analysis} channel for deep, step-by-step reasoning and scratchpad.}

\texttt{- Use the \textbf{final} channel to output a clear, concrete \textbf{plan} for the next actions.}

\texttt{- Do \textbf{not} actually call tools in this turn; only describe how the tools should be used in the future steps.}
\end{quote}

The developer prompt provides additional guidance on plan quality:

\begin{quote}
\small
\texttt{Planner reasoning guidance:}
\begin{itemize}[noitemsep,topsep=0pt]
\item \texttt{Think \textbf{step by step} and \textbf{deeply} in your own natural style.}
\item \texttt{Focus on the \textbf{next several actions} and how to verify progress on the page.}
\item \texttt{Ground your plan in: the \textbf{user task}, prior \textbf{plans}, executed \textbf{tool calls} and their results, the latest \textbf{HTML}.}
\item \texttt{Avoid unsupported assumptions about elements or states that you have not observed.}
\end{itemize}

\texttt{Recovery and loop-avoidance:}
\begin{itemize}[noitemsep,topsep=0pt]
\item \texttt{Do \textbf{not} repeat plans or actions that failed or produced no state change.}
\item \texttt{When something did not work, explicitly include \textbf{recovery steps}: different id, scroll, search, hover, go\_back, or another sensible control.}
\item \texttt{Clarify how the Executor should adapt if the page does not look as expected.}
\end{itemize}

\texttt{Referencing future actions:}
\begin{itemize}[noitemsep,topsep=0pt]
\item \texttt{Refer to tools by their names (for example: click, type\_text, hover, scroll, select\_dropdown\_option, search, go\_back, exit), as they appear in the tool schemas.}
\item \texttt{Describe \textbf{how} the Executor should pick the target element: use the integer id from the latest HTML when known, otherwise reference robust cues (labels, nearby text, attributes, structural context).}
\item \texttt{Predict how the \textbf{DOM or URL should change} after key actions.}
\item \texttt{Specify what the Executor should look for to \textbf{confirm success vs. failure} (new headings, result lists, modals, disappearance of buttons, etc.).}
\end{itemize}
\end{quote}

\section{Benchmark Characteristics and LLM-as-Judge Reliability}
\label{app:benchmark_characteristics}

\paragraph{Benchmark difficulty comparison.}
Our three evaluation benchmarks differ substantially in task difficulty, trajectory length, and evaluation methodology:

\begin{itemize}
  \item \textbf{WebArena-Lite} (165 tasks): Uses programmatic success checks that verify specific page states or database entries. Tasks often require longer trajectories with precise multi-step reasoning. Average trajectory length is approximately 8-12 steps, and success rates are generally lower (40-47\% in our experiments).

  \item \textbf{Online-Mind2Web} (300 tasks): Draws from 136 diverse real-world websites (e-commerce, travel, government, etc.) using the Online-Mind2Web dataset \citep{xue2025an}. It uses an LLM-as-judge evaluation protocol with \JUDGEMODEL{}. Also, it features the longest trajectories among our benchmarks (mean 17.4 steps), making it a particularly challenging testbed for long-horizon scaling. Success rates range from 15-32\% in our experiments, reflecting both the difficulty of the tasks and the diversity of the web environments.

  \item \textbf{GoBrowse-style} (341 tasks): Uses an LLM-as-judge evaluation protocol. Tasks tend to be shorter and more straightforward, with average trajectory lengths of 4-6 steps. Success rates are substantially higher (86-90\% in our experiments), reflecting easier task specifications.
\end{itemize}

The three benchmarks span a wide range of difficulty levels: GoBrowse (86--90\% baseline) has less room for improvement, Online-Mind2Web (15--22\% baseline) provides the most headroom, and WebArena-Lite (38--43\% baseline) sits in between. Nevertheless, \OURS achieves consistent gains on all three benchmarks, suggesting the underlying principles generalize across difficulty levels.

\paragraph{LLM-as-judge reliability.}
Both Online-Mind2Web and GoBrowse use LLM-as-judge evaluation. For GoBrowse, we employ the evaluation protocol introduced in the original work; based on validation studies reported there, the judge achieves approximately 90\% agreement with human evaluations on task success determination. For Online-Mind2Web, we use \JUDGEMODEL{} as the judge model, following the WebJudge protocol that scores trajectories based on key action points extracted from task descriptions and final screenshots.

While LLM-as-judge evaluation introduces some noise compared to programmatic checks, the consistent improvement pattern observed across all three benchmarks (WebArena-Lite with programmatic evaluation, Online-Mind2Web and GoBrowse with LLM-as-judge) provides confidence in our findings.

\section{Semantic Deduplication Ablation}
\label{app:deduplication_ablation}

\tref{tab:dedup_ablation} demonstrates the importance of semantic deduplication for majority voting. Without deduplication (``Before fixes''), increasing the number of candidates $N$ can \emph{hurt} performance on GoBrowse: accuracy drops from 83.3\% at $N{=}1$ to 80.1\% at $N{=}32$. This counterintuitive result occurs because semantically equivalent actions (e.g., ``N/A'' vs.\ ``Not found'', or minor rephrasing of search queries) split votes, causing the majority vote to select incorrect minority actions.

After applying semantic deduplication (``After fixes''), scaling behavior normalizes and performance improves. For ReAct at $N{=}8$, accuracy increases from 83.3\% to 84.5\%. The per-website breakdown shows consistent improvements across domains, with particularly large gains on Reddit (84.8\%$\rightarrow$94.9\%) where text input variations are common.

\begin{table}[h]
  \centering
  \caption{\textbf{Impact of semantic deduplication on GoBrowse.} Without deduplication, majority voting degrades with more samples due to vote splitting among semantically equivalent actions. With deduplication, expected scaling behavior is restored and performance improves. This ablation confirms that semantic deduplication is a necessary preprocessing step for meaningful vote aggregation, not a confounding factor.}
  \label{tab:dedup_ablation}
  \vspace{1mm}
  \small
  \begin{tabular}{@{}llcccccc@{}}
    \toprule
    & & \multicolumn{6}{c}{After Deduplication (per-website)} \\
    \cmidrule(l){3-8}
    Setting & Before & Overall & GitLab & Map & Reddit & Shopping & Shop.Admin \\
    \midrule
    \multicolumn{8}{l}{\textit{ReAct}} \\
    $N{=}1$ & 83.3\% & 83.3\% & 84.7\% & 76.3\% & 84.8\% & 89.8\% & 82.7\% \\
    $N{=}8$ & 83.3\% & 84.5\% & 81.9\% & 77.6\% & 94.9\% & 86.4\% & 84.0\% \\
    $N{=}16$ & 81.8\% & 82.1\% & 83.3\% & 73.7\% & 94.8\% & 88.1\% & 74.7\% \\
    $N{=}32$ & 80.1\% & -- & -- & -- & -- & -- & -- \\
    \midrule
    \multicolumn{8}{l}{\textit{Plan\&Act}} \\
    $(1,1)$ & 82.4\% & 82.4\% & -- & -- & -- & -- & -- \\
    $(2,4)$ & 81.2\% & 81.2\% & -- & -- & -- & -- & -- \\
    $(4,2)$ & 84.2\% & 85.3\% & 87.5\% & 77.6\% & 91.4\% & 91.5\% & 81.3\% \\
    \bottomrule
  \end{tabular}
\end{table}

This ablation addresses a potential concern that semantic deduplication artificially inflates results. On the contrary, the data shows that deduplication is \emph{required} for majority voting to function as intended. Without it, the method exhibits behavior where more samples lead to worse outcomes. The improvement from deduplication (e.g., 80.1\%$\rightarrow$84.5\% at $N{=}8$) reflects the restoration of correct voting semantics, and is not an artificial boost.

\section{Plan-and-Act Scaling Results}
\label{app:plan_and_act}

We evaluated Plan-and-Act agents \citep{erdogan2025plan} with factorized candidate-generation budgets to determine whether non-monotonic scaling is specific to ReAct-style agents or a more general phenomenon.
Let $P$ denote the number of plan candidates sampled and $A$ the number of action candidates per plan, yielding total budget $C = P \times A$.
\tref{tab:plan_and_act_full} presents results on GoBrowse.

\begin{table}[h]
  \centering
  \caption{\textbf{Plan-and-Act scaling on WebArena-Lite and GoBrowse.} Success rates under different budget allocations $(P, A)$ with total compute $C = P \times A$. Non-monotonic scaling is evident on both benchmarks: on WebArena-Lite, $(2,4)$ achieves the best result (43.2\%) but $(4,4)$ drops to 43.0\% despite 2$\times$ the budget; on GoBrowse, the baseline $(1,1)$ achieves 83.3\%, while $(2,4)$ with $8\times$ the budget drops to 80.6\%. This confirms that the non-monotonic scaling phenomenon affects multiple agent architectures, not just ReAct, strengthening the motivation for dynamic compute allocation.}
  \label{tab:plan_and_act_full}
  \vspace{1mm}
  \small
  \begin{tabular}{@{}lcccc@{}}
    \toprule
    $(P, A)$ & Budget $C$ & WA & GB & Tokens \\
    \midrule
    $(1,1)$ & 1 & 38.8\% & 83.3\% & 104K \\
    $(2,4)$ & 8 & \textbf{43.2\%} & 80.6\% & 731K \\
    $(4,4)$ & 16 & 43.0\% & 81.5\% & 1.4M \\
    \bottomrule
  \end{tabular}
\end{table}

The results mirror what we observe with ReAct: scaling up compute via more samples does not reliably improve performance.
In fact, performance \emph{decreases} from 83.3\% to 80.6\% when moving from budget $C{=}1$ to $C{=}8$.
This establishes that the non-monotonic scaling phenomenon affects multiple agent architectures, providing additional motivation for the dynamic compute allocation approach developed in \sref{sec:adis}.

\section{Complete Arbiter Scaling Results}
\label{app:arbiter_scaling_full}

\tref{tab:arbiter_scaling_full} presents the complete arbiter scaling results on WebArena-Lite across all combinations of candidate count $N \in \{3, 5, 10, 20\}$ and arbiter scaling factor $K \in \{1, 5, 10, 20\}$. Success rates are averaged over 3 independent runs.

\begin{table}[h]
  \centering
  \caption{\textbf{Complete arbiter scaling on WebArena-Lite.} Success rates for all $(N, K)$ configurations, averaged over 3 runs. Best result for each $N$ is bolded. The optimal arbiter scaling factor $K$ varies by candidate count $N$: small $N$ benefits from high $K$ ($N{=}3$, $K{=}20$: 45.7\%), while larger $N$ prefers moderate $K$ ($N{=}10$, $K{=}5$: 44.6\%). No single $(N,K)$ dominates, and gains from arbiter scaling plateau or decline beyond optimal $K$. This non-monotonic behavior motivates selective arbiter use via dynamic gating rather than uniform high-$K$ allocation.}
  \label{tab:arbiter_scaling_full}
  \vspace{1mm}
  \small
  \begin{tabular}{@{}ccccc@{}}
    \toprule
    & \multicolumn{4}{c}{Arbiter Scaling Factor $K$} \\
    \cmidrule(l){2-5}
    Candidates $N$ & $K{=}1$ & $K{=}5$ & $K{=}10$ & $K{=}20$ \\
    \midrule
    $N{=}3$ & 43.0\% & 41.2\% & 43.4\% & \textbf{45.7\%} \\
    $N{=}5$ & 42.8\% & 44.2\% & \textbf{44.6\%} & 42.0\% \\
    $N{=}10$ & 44.0\% & \textbf{44.6\%} & 42.0\% & 43.0\% \\
    $N{=}20$ & 43.2\% & 42.8\% & \textbf{44.0\%} & 43.6\% \\
    \bottomrule
  \end{tabular}
\end{table}

\section{RSA and PlanRSA Full Results}
\label{app:rsa_full}

We evaluated Recursive Self-Aggregation (RSA) \citep{rsa2025} as an alternative to single-round arbitration. RSA iteratively refines candidates over $T$ rounds, with $K$ aggregation calls per round. 
Building on top of Plan-and-Act, we combine it with RSA to create PlanRSA, where we apply RSA at the plan level.
\tref{tab:rsa_full} presents results across all configurations tested on WebArena-Lite.

\begin{table}[ht]
  \centering
  \caption{\textbf{Full RSA and PlanRSA results on WebArena-Lite.} RSA applies iterative refinement at the action level, while PlanRSA applies it at the plan level for Plan-and-Act agents. Despite substantially higher compute costs (up to 80 LLM calls per step), RSA achieves at best comparable performance to single-round arbitration. PlanRSA underperforms all baselines, suggesting that aggregation noise accumulates more severely at higher levels of abstraction.}
  \label{tab:rsa_full}
  \vspace{1mm}
  \small
  \begin{tabular}{@{}lccccc@{}}
    \toprule
    Method & $N/P$ & $K$ & $T$ & Success & Calls/Step \\
    \midrule
    \multicolumn{6}{l}{\textit{Baselines}} \\
    Majority Vote & 10 & -- & -- & 43.2\% & 10 \\
    Arbiter ($K{=}1$) & 10 & 1 & -- & 44.0\% & 11 \\
    \OURS (entropy, best) & 10 & 1 & -- & 47.9\% & $\sim$7 \\
    \midrule
    \multicolumn{6}{l}{\textit{RSA (action-level aggregation)}} \\
    RSA & 8 & 2 & 2 & 41.0\% & 24 \\
    RSA & 8 & 2 & 4 & 41.0\% & 40 \\
    RSA & 16 & 4 & 2 & 42.0\% & 48 \\
    RSA & 16 & 4 & 4 & 43.6\% & 80 \\
    \midrule
    \multicolumn{6}{l}{\textit{PlanRSA (plan-level aggregation)}} \\
    PlanRSA (arbiter select) & 8 & 4 & 2 & 35.2\% & 40 \\
    PlanRSA (random select) & 8 & 4 & 2 & 35.2\% & 40 \\
    \bottomrule
  \end{tabular}
\end{table}

\paragraph{Analysis.}
RSA with $N{=}16, K{=}4, T{=}4$ achieves 43.6\%, comparable to single-round arbitration (44.0\%) but at ${\sim}7\times$ the compute cost per step. The marginal benefit of additional aggregation rounds is minimal: increasing $T$ from 2 to 4 improves performance by only 0--1.6\% while roughly doubling compute. We hypothesize that RSA's iterative refinement, designed for single-shot reasoning tasks where solutions can be progressively improved, does not transfer well to per-step action selection where the ``correct'' answer depends heavily on environmental context that cannot be improved through reasoning alone.

PlanRSA performs substantially worse than all baselines (35.2\% vs.\ 43.2\% for majority vote). Notably, the selection method within PlanRSA (arbiter vs.\ random) makes no difference, suggesting the bottleneck is the plan aggregation process itself rather than final selection. This may reflect error accumulation: aggregating plans introduces noise at a higher level of abstraction, and errors in plan formulation propagate to all downstream actions.

\section{Threshold Sensitivity Analysis}
\label{app:threshold_sensitivity}

Tables~\ref{tab:entropy_sweep} and~\ref{tab:margin_sweep} present the full threshold sweep results for \OURS on WebArena-Lite across different values of $N \in \{3, 5, 10, 20\}$. Both entropy-gated and margin-gated variants show robust performance across a range of thresholds $\tau \in \{0.2, 0.3, \ldots, 0.8\}$. The best-performing threshold for each $N$ is highlighted in bold.

\begin{table}[h]
  \centering
  \caption{\textbf{\OURS threshold sweep on WebArena-Lite.} Success rates across different candidate counts ($N$) and threshold values ($\tau$) for entropy-gated (left) and margin-gated (right) variants. Best-performing threshold for each $N$ is bolded. Baseline majority vote success rates: $N{=}3$: 40.2\%, $N{=}5$: 42.8\%, $N{=}10$: 43.2\%, $N{=}20$: 43.8\%. Most configurations outperform their respective baselines, demonstrating robustness to threshold choice.}
  \label{tab:threshold_sweep}
  \vspace{1mm}
  \small
  \begin{subtable}[t]{0.48\linewidth}
    \centering
    \caption{Entropy-gated}
    \label{tab:entropy_sweep}
    \begin{tabular}{@{}ccccc@{}}
      \toprule
      $\tau$ & $N{=}3$ & $N{=}5$ & $N{=}10$ & $N{=}20$ \\
      \midrule
      0.2 & 44.2 & 43.0 & \textbf{47.9} & 43.6 \\
      0.3 & 43.0 & 44.2 & 46.7 & 44.2 \\
      0.4 & \textbf{46.7} & 45.5 & 44.2 & 46.1 \\
      0.5 & 43.0 & 45.5 & 45.5 & 40.6 \\
      0.6 & 44.2 & 45.5 & 46.1 & 43.0 \\
      0.7 & 44.2 & 44.2 & 47.3 & \textbf{47.3} \\
      0.8 & 42.4 & 45.5 & 45.5 & 42.4 \\
      \bottomrule
    \end{tabular}
  \end{subtable}
  \hfill
  \begin{subtable}[t]{0.48\linewidth}
    \centering
    \caption{Margin-gated}
    \label{tab:margin_sweep}
    \begin{tabular}{@{}ccccc@{}}
      \toprule
      $\tau$ & $N{=}3$ & $N{=}5$ & $N{=}10$ & $N{=}20$ \\
      \midrule
      0.2 & 44.8 & 45.5 & 45.5 & 44.8 \\
      0.3 & 43.6 & 45.5 & 42.4 & 46.7 \\
      0.4 & 46.1 & 45.5 & \textbf{46.1} & 44.2 \\
      0.5 & 43.0 & \textbf{47.9} & 44.8 & 44.2 \\
      0.6 & 43.6 & 44.2 & 44.2 & 46.7 \\
      0.7 & 44.2 & 44.2 & 44.2 & \textbf{47.9} \\
      0.8 & \textbf{47.3} & 46.1 & \textbf{46.1} & 44.2 \\
      \bottomrule
    \end{tabular}
  \end{subtable}
\end{table}

The results demonstrate that \OURS is relatively robust to the choice of threshold across different values of $N$. For entropy-gated \OURS, the best thresholds vary by $N$ but most configurations outperform the majority vote baseline. For margin-gated \OURS, similar robustness is observed with peak performance typically at $\tau \in \{0.4, 0.5, 0.7, 0.8\}$. In practice, a default threshold of $\tau = 0.5$ provides good performance across both benchmarks without extensive tuning.

\section{Threshold Calibration from Limited Data}
\label{app:threshold_calibration}

A natural practical question is: \emph{how should a practitioner select $\tau$ when deploying \OURS in a new setting?}
We address this through comprehensive bootstrap calibration experiments ($R{=}200$ stratified rounds) across WebArena-Lite (165 tasks, 5 websites) and Online-Mind2Web (300 tasks), using both entropy-based and margin-based gating at $N{=}10$.

\subsection{Methodology}

We use a stratified bootstrap procedure to simulate realistic calibration:
\begin{enumerate}[nosep]
    \item \textbf{Split.} Randomly draw $X\%$ of tasks (stratified by website where applicable) as the calibration set.
    \item \textbf{Calibrate.} On the calibration set, evaluate each candidate $\tau \in \{0.2, 0.3, \ldots, 0.8\}$ and select $\hat{\tau} = \arg\max_\tau \text{acc}_\text{cal}(\tau)$.
    \item \textbf{Evaluate.} Apply $\hat{\tau}$ to the \emph{entire benchmark} and measure $\text{acc}_\text{full}(\hat{\tau})$.
    \item \textbf{Oracle comparison.} The oracle threshold $\tau^* = \arg\max_\tau \text{acc}_\text{full}(\tau)$ is computed on the full benchmark.
    \item \textbf{Regret.} $\text{regret} = \text{acc}_\text{full}(\tau^*) - \text{acc}_\text{full}(\hat{\tau})$.
\end{enumerate}
We repeat this $R{=}200$ times for each combination of calibration fraction $X \in \{10\%, 15\%, 20\%, 25\%, 30\%, 40\%, 50\%\}$ and gating method (entropy, margin).
The \textbf{regret} measures how much accuracy is lost by using the calibrated $\hat{\tau}$ instead of the full-benchmark oracle.

\subsection{Universal Calibration Results}

In \emph{universal calibration}, we select a single $\hat{\tau}$ that applies to all tasks.
\tref{tab:calibration_universal} presents results on both benchmarks.

\begin{table}[h]
  \centering
  \caption{\textbf{Universal calibration results ($N{=}10$, 200 bootstrap rounds).} ``Agreement'' is the fraction of rounds where $\hat{\tau}$ matches the oracle $\tau^*$. ``Regret'' is the mean accuracy gap (\%). Even at $X{=}10\%$ (${\sim}16$ tasks on WA-Lite), regret is ${\leq}1.1\%$ across all settings. Best results (at $X{=}50\%$) are highlighted.}
  \label{tab:calibration_universal}
  \small
  \begin{tabular}{@{}l cccc cccc@{}}
    \toprule
    & \multicolumn{4}{c}{\textbf{WebArena-Lite}} & \multicolumn{4}{c}{\textbf{Online-Mind2Web}} \\
    \cmidrule(lr){2-5} \cmidrule(lr){6-9}
    & \multicolumn{2}{c}{Entropy} & \multicolumn{2}{c}{Margin} & \multicolumn{2}{c}{Entropy} & \multicolumn{2}{c}{Margin} \\
    \cmidrule(lr){2-3} \cmidrule(lr){4-5} \cmidrule(lr){6-7} \cmidrule(lr){8-9}
    $X$ & Agree. & Regret & Agree. & Regret & Agree. & Regret & Agree. & Regret \\
    \midrule
    10\% & 42\% & 0.99\% & 18\% & 1.01\% & 18\% & 1.08\% & 16\% & 1.81\% \\
    15\% & 35\% & 1.12\% & 24\% & 0.98\% & 26\% & 0.76\% & 14\% & 1.63\% \\
    20\% & 40\% & 0.99\% & 22\% & 0.98\% & 22\% & 0.92\% & 17\% & 1.12\% \\
    25\% & 36\% & 1.08\% & 26\% & 0.87\% & 29\% & 0.63\% & 19\% & 1.23\% \\
    30\% & 40\% & 0.92\% & 25\% & 0.77\% & 22\% & 0.69\% & 28\% & 0.81\% \\
    40\% & 44\% & 0.91\% & 27\% & 0.66\% & 26\% & 0.59\% & 18\% & 0.88\% \\
    \rowcolor{green!12}
    50\% & 46\% & \textbf{0.77\%} & 28\% & \textbf{0.55\%} & 25\% & \textbf{0.49\%} & 35\% & \textbf{0.56\%} \\
    \bottomrule
  \end{tabular}
\end{table}

Several patterns emerge:
(1)~\textbf{Regret is remarkably small.} Even at $X{=}10\%$ (${\sim}16$ tasks on WebArena-Lite, ${\sim}30$ tasks on Mind2Web), regret is ${\leq}1.1\%$ for entropy gating across both benchmarks. Since \OURS typically improves over majority voting by 3--12\%, losing ${\sim}1\%$ to imperfect calibration preserves the vast majority of the benefit.
(2)~\textbf{Regret decreases with more data.} WA-Lite margin regret decreases from 1.01\% to 0.55\% as $X$ grows from 10\% to 50\%.
(3)~\textbf{Low agreement does not imply high regret.} Agreement rates are moderate (25--46\% at $X{=}50\%$), yet regret is minimal because the accuracy landscape is flat near the optimum.

\subsection{Why Calibration Works: Diagnostic Analysis}

The coexistence of moderate agreement and very low regret is explained by the \emph{flat accuracy landscape} near the optimum.
The threshold selection variance $\sigma_\tau$ remains around 0.17--0.22 across all calibration fractions, confirming that the calibration procedure does not converge to a single threshold even with abundant data, because \textbf{multiple thresholds are genuinely near-optimal}.
For example, on WebArena-Lite entropy ($N{=}10$), the oracle is $\tau{=}0.2$ (47.9\%), but $\tau{=}0.3$ achieves 46.7\% and $\tau{=}0.7$ achieves 47.3\%, so selecting any of these ``wrong'' thresholds costs at most 1.2\%.

\begin{figure}[h]
    \centering
    \includegraphics[width=0.6\linewidth]{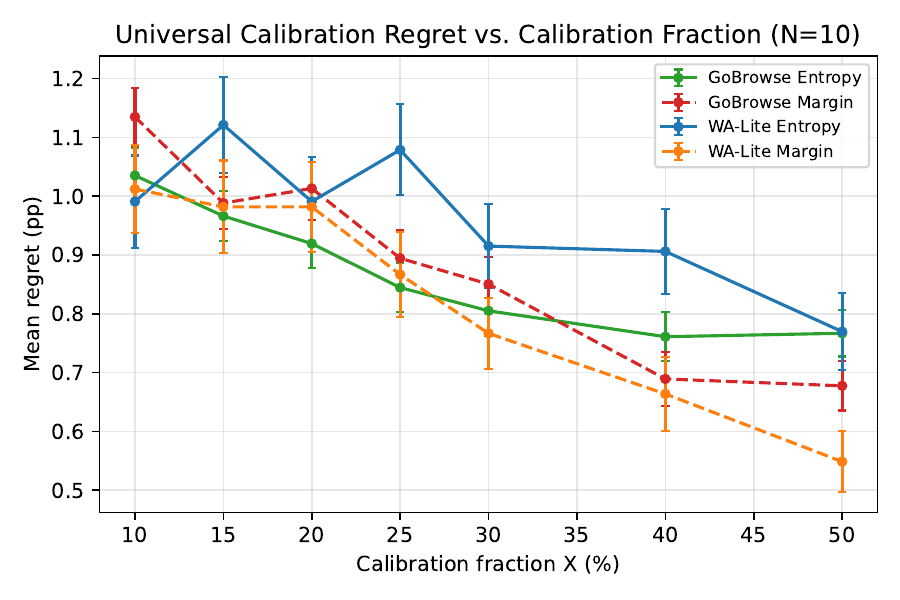}
    \caption{\textbf{Calibration regret vs.\ calibration fraction} ($N{=}10$, 200 rounds). All curves start at ${\leq}1.1\%$ even at $X{=}10\%$ and decrease monotonically. 15--20\% of tasks suffices for effective calibration.}
    \label{fig:regret_curve}
\end{figure}

\fref{fig:regret_curve} provides a direct visualization of how regret decreases with more calibration data across both benchmarks.

\subsection{Per-Website Threshold Calibration}

Universal calibration selects a single $\hat{\tau}$ for all tasks. \emph{Per-website calibration} selects a separate $\hat{\tau}_w$ for each website $w$, gaining flexibility but losing statistical power (smaller per-website subsets).

\tref{tab:pw_calibration} compares universal and per-website calibration for both gating methods. Notably, per-website calibration can achieve \emph{negative} regret, meaning it \emph{exceeds} the global oracle accuracy by exploiting website-specific threshold preferences.

\begin{table}[h]
\centering
\caption{\textbf{Universal vs.\ per-website calibration ($N{=}10$).} \colorbox{green!15}{Green cells} indicate per-website regret that is negative (exceeds the global oracle). ``PW Adv.'' is the advantage of per-website over universal calibration. All values in \%.}
\label{tab:pw_calibration}
\small
\begin{subtable}[t]{\linewidth}
\centering
\caption{Entropy gating}
\begin{tabular}{@{}l ccc ccc@{}}
\toprule
& \multicolumn{3}{c}{\textbf{WebArena-Lite}} & \multicolumn{3}{c}{\textbf{GoBrowse}} \\
\cmidrule(lr){2-4} \cmidrule(lr){5-7}
$X$ & Univ. & PW Regret & PW Adv. & Univ. & PW Regret & PW Adv. \\
\midrule
10\% & 0.99 & \cellcolor{green!15}\textbf{$-$0.26} & \textcolor{forestgreen}{+1.25} & 1.04 & 0.44 & \textcolor{forestgreen}{+0.60} \\
20\% & 0.99 & \cellcolor{green!15}\textbf{$-$0.61} & \textcolor{forestgreen}{+1.60} & 0.92 & \cellcolor{green!15}\textbf{$-$0.26} & \textcolor{forestgreen}{+1.18} \\
30\% & 0.92 & \cellcolor{green!20}\textbf{$-$1.23} & \textcolor{forestgreen}{+2.15} & 0.80 & \cellcolor{green!15}\textbf{$-$0.63} & \textcolor{forestgreen}{+1.44} \\
50\% & 0.77 & \cellcolor{green!25}\textbf{$-$2.14} & \textcolor{forestgreen}{+2.91} & 0.77 & \cellcolor{green!20}\textbf{$-$1.41} & \textcolor{forestgreen}{+2.18} \\
\bottomrule
\end{tabular}
\end{subtable}

\vspace{1em}

\begin{subtable}[t]{\linewidth}
\centering
\caption{Margin gating}
\begin{tabular}{@{}l ccc ccc@{}}
\toprule
& \multicolumn{3}{c}{\textbf{WebArena-Lite}} & \multicolumn{3}{c}{\textbf{GoBrowse}} \\
\cmidrule(lr){2-4} \cmidrule(lr){5-7}
$X$ & Univ. & PW Regret & PW Adv. & Univ. & PW Regret & PW Adv. \\
\midrule
10\% & 1.01 & 0.05 & \textcolor{forestgreen}{+0.97} & 1.13 & 0.41 & \textcolor{forestgreen}{+0.73} \\
20\% & 0.98 & \cellcolor{green!15}\textbf{$-$0.21} & \textcolor{forestgreen}{+1.19} & 1.01 & \cellcolor{green!15}\textbf{$-$0.27} & \textcolor{forestgreen}{+1.28} \\
30\% & 0.77 & \cellcolor{green!15}\textbf{$-$0.77} & \textcolor{forestgreen}{+1.54} & 0.85 & \cellcolor{green!15}\textbf{$-$0.52} & \textcolor{forestgreen}{+1.37} \\
50\% & 0.55 & \cellcolor{green!20}\textbf{$-$1.60} & \textcolor{forestgreen}{+2.15} & 0.68 & \cellcolor{green!20}\textbf{$-$1.24} & \textcolor{forestgreen}{+1.91} \\
\bottomrule
\end{tabular}
\end{subtable}
\end{table}

\begin{figure}[h]
    \centering
    \begin{subfigure}[t]{0.48\linewidth}
        \centering
        \includegraphics[width=\linewidth]{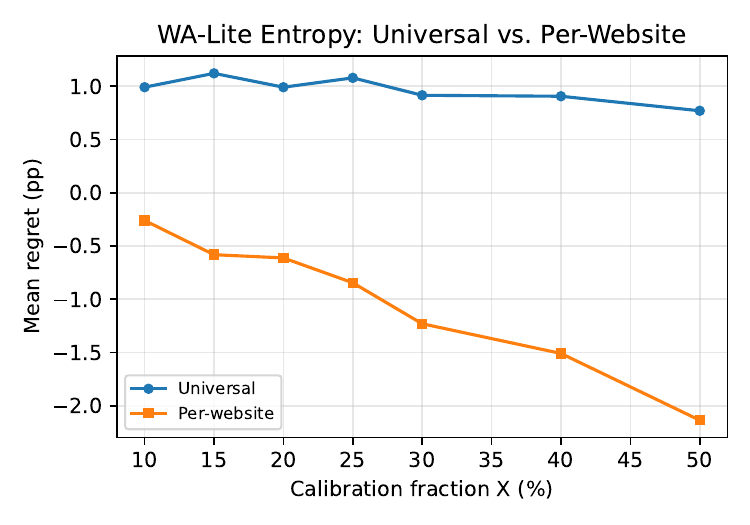}
        \caption{WA-Lite, entropy.}
    \end{subfigure}
    \hfill
    \begin{subfigure}[t]{0.48\linewidth}
        \centering
        \includegraphics[width=\linewidth]{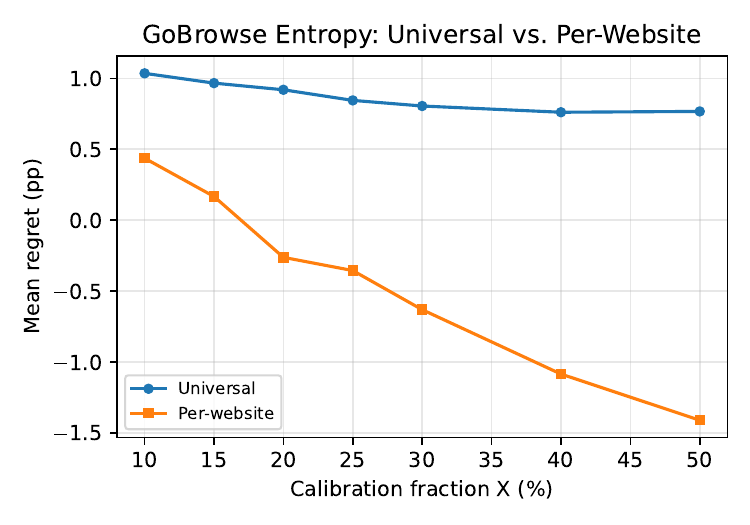}
        \caption{GoBrowse, entropy.}
    \end{subfigure}
    \\[0.5em]
    \begin{subfigure}[t]{0.48\linewidth}
        \centering
        \includegraphics[width=\linewidth]{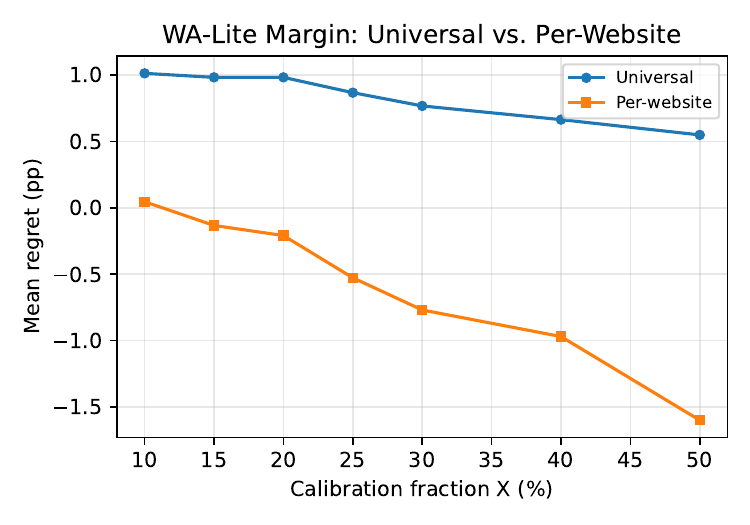}
        \caption{WA-Lite, margin.}
    \end{subfigure}
    \hfill
    \begin{subfigure}[t]{0.48\linewidth}
        \centering
        \includegraphics[width=\linewidth]{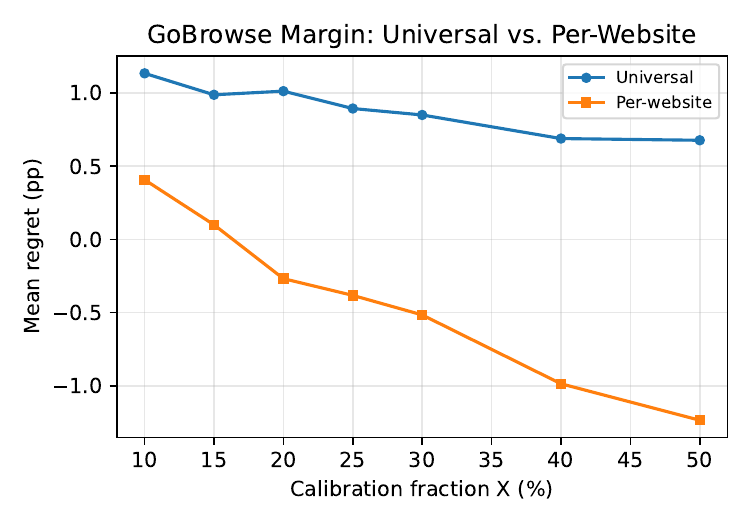}
        \caption{GoBrowse, margin.}
    \end{subfigure}
    \caption{\textbf{Universal vs.\ per-website calibration regret.} Per-website (orange) consistently achieves lower regret than universal (blue), with the gap widening at larger $X$. Per-website regret frequently turns negative, meaning it exceeds the global oracle.}
    \label{fig:universal_vs_pw}
\end{figure}

\fref{fig:universal_vs_pw} visualizes the comparison across all settings.

\subsection{Summary and Practical Guidance}

\begin{enumerate}[leftmargin=*,nosep]
    \item \textbf{Universal calibration is remarkably effective.} With as little as 10\% of tasks, regret is ${\leq}1.1\%$ across all settings. The flat accuracy landscape means even imprecise threshold selection works.
    \item \textbf{Per-website calibration exceeds the global oracle.} At $X{\geq}15$--$20\%$, per-website regret turns negative, exploiting website-specific threshold preferences. The advantage grows with $X$, reaching $+2.91\%$ for WA-Lite entropy at $X{=}50\%$.
    \item \textbf{Practical recommendation:}
    (a)~With no data, use $\tau{=}0.5$ as a robust default;
    (b)~With 10--20\% of tasks, run universal calibration (${\leq}1.1\%$ regret);
    (c)~With 20\%+ tasks and website labels, use per-website calibration to exceed the oracle.
\end{enumerate}

\section{Vote Distribution Analysis}
\label{app:vote_distribution}

\begin{figure}[h]
  \centering
  \includegraphics[width=\linewidth]{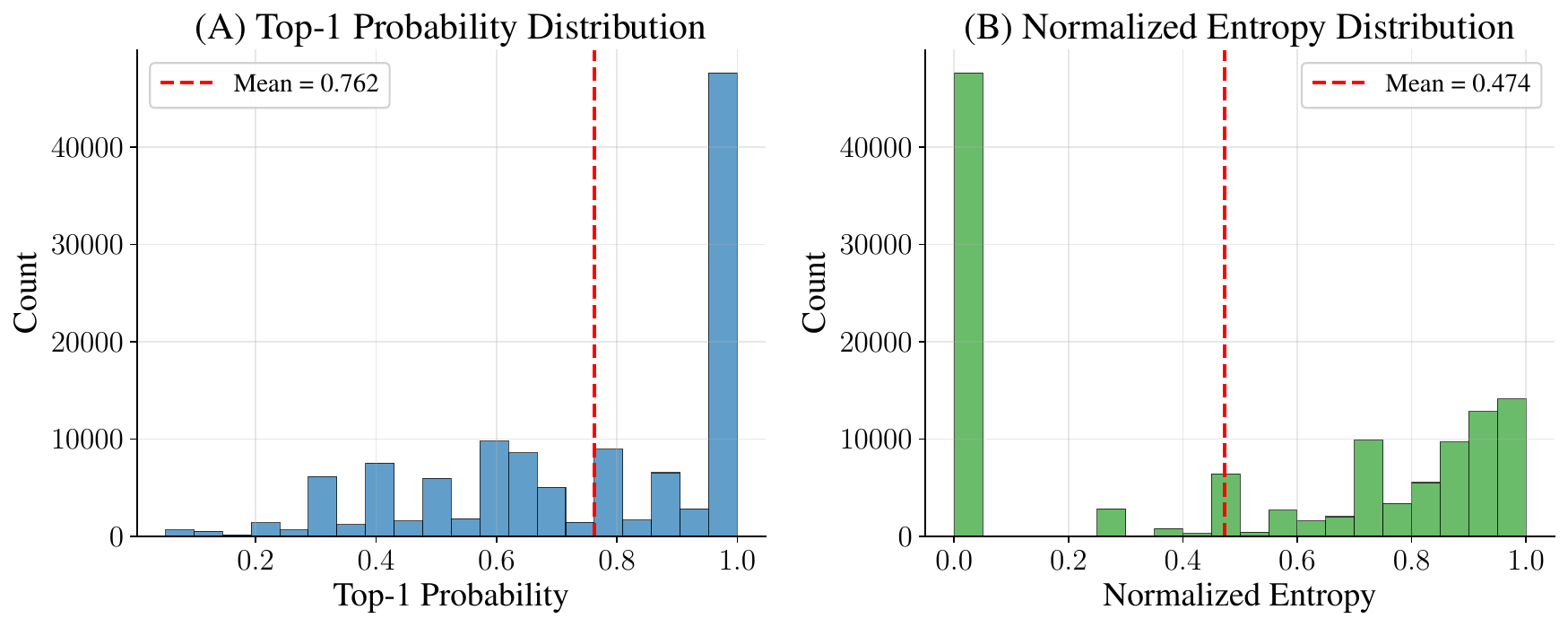}
  \caption{\textbf{Vote distribution profiles across all steps.} Histograms aggregated over all decision steps from $N{=}10$ experiments on WebArena-Lite. (A) Top-1 probability distribution shows strong right skew: ${\sim}42\%$ of steps have near-deterministic consensus (top-1 probability ${>}0.9$), and mean top-1 probability is 0.762. This indicates many steps fall into a redundancy regime where additional candidates duplicate the dominant action. (B) Normalized entropy distribution is bimodal (mean 0.474): ${\sim}40\%$ of steps have zero entropy (perfect consensus), while ${\sim}49\%$ of steps have entropy ${>}0.6$, identifying contentious steps where arbitration is most valuable. These profiles motivate dynamic gating: allocate extra compute to high-entropy steps while preserving consensus elsewhere.}
  \label{fig:consensus_profile}
\end{figure}

\fref{fig:consensus_profile} presents the detailed analysis of vote distributions across all decision steps from our experiments. This analysis provides empirical support for the two-regime interpretation discussed in \sref{sec:results}.

The distributions reveal several key patterns: (1) \textbf{Right-skewed top-1 probability:} Approximately 42\% of steps exhibit near-deterministic consensus with top-1 probability exceeding 0.9, indicating that many decisions fall into a redundant regime. (2) \textbf{Bimodal entropy:} The normalized entropy distribution is bimodal (mean 0.474), with approximately 40\% of steps at zero entropy (perfect consensus) and approximately 49\% showing entropy above 0.6. (3) \textbf{Two-regime interpretation:} The distributions strongly support the two-regime interpretation from \sref{sec:results}: a large fraction of steps are routine (high consensus) while a comparable fraction are contentious (genuine uncertainty), motivating dynamic compute allocation.

These empirical profiles directly motivate the dynamic gating approach in \OURS: by allocating extra arbitration compute to the high-entropy tail while preserving consensus on the dominant low-entropy steps, we can improve the accuracy--compute tradeoff without the overhead of uniform scaling.

\section{Failure Mode Examples}
\label{app:failure_examples}

This section provides a concrete example from our experiments where the arbiter's override of high-consensus decisions led to task failure. This example illustrates the core failure mode discussed in \sref{sec:uncertainty_diagnostic}: when the vote distribution already exhibits strong agreement (high margin $\Delta_t$, low entropy $H_t$), arbitration introduces override risk rather than improving decision quality.

\begin{figure}[t]
  \centering
  \includegraphics[width=0.7\columnwidth, trim=150 150 170 150, clip]{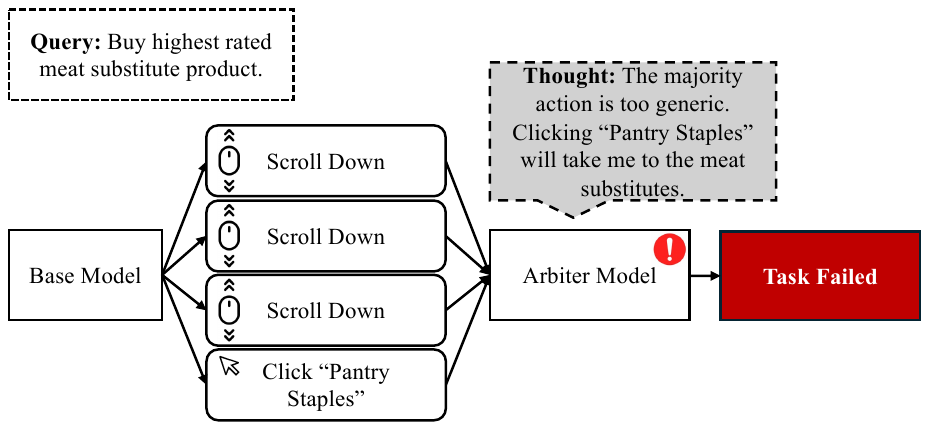}
  \caption{\textbf{Arbiter failure on high-consensus step.} Example trajectory where the sampled candidates exhibit strong consensus (e.g., 9/10 votes for one action), but the arbiter overrides the majority and selects a minority action, leading to task failure. This illustrates that arbitration can be harmful when vote-derived uncertainty is already low.}
  \label{fig:arbiter_failure}
\end{figure}

Figure~\ref{fig:arbiter_failure} presents an example from a WebArena-Lite task requiring the agent to find ``Meat Substitutes'' in an online grocery store. At a critical decision point, the sampled candidates show 90\% consensus for ``Scroll Down'', the correct action, since the target category lies below the current viewport. However, the arbiter overrides this strong consensus and selects ``Click Pantry Staples,'' a minority action that navigates to an incorrect category and derails the trajectory.

This failure pattern is particularly instructive because the vote distribution provides a clear signal: with 9 out of 10 candidates agreeing on ``Scroll,'' the normalized entropy is low ($H_t \approx 0.3$) and the margin is high ($\Delta_t = 0.8$). Under \OURS, this step would be classified as high-confidence and the arbiter would be bypassed entirely, preserving the consensus action. The example thus demonstrates both (i) why uniform arbitration can be harmful, and (ii) how vote-derived uncertainty enables selective intervention.

\section{DeepConf Full Results}
\label{app:deepconf_full}

\tref{tab:deepconf_results} presents the full DeepConf results across three confidence variants and varying candidate counts $N$ on both benchmarks.

\begin{table}[h]
  \centering
  \caption{\textbf{DeepConf variants on WebArena-Lite and Online-Mind2Web.} Unlike \OURS, DeepConf requires access to token-level log probabilities.}
  \label{tab:deepconf_results}
  \small
\begin{tabular}{@{}llcccc@{}}
    \toprule
    & & \multicolumn{2}{c}{WebArena-Lite} & \multicolumn{2}{c}{Online-Mind2Web} \\
    \cmidrule(lr){3-4} \cmidrule(lr){5-6}
    Variant & $N$ & Success & Tokens & Success & Tokens \\
    \midrule
    \multirow{3}{*}{Avg.\ Trace}
      & 3  & 40.2\% & 262K & 24.7\% & 465K \\
      & 5  & 42.0\% & 408K & 20.7\% & 834K \\
      & 10  & \textbf{43.8\%} & 828K & 20.0\% & 1.7M \\
    \midrule
    \multirow{3}{*}{Tail}
      & 3  & 38.8\% & 252K & 22.3\% & 458K \\
      & 5  & 40.4\% & 408K & 19.3\% & 820K \\
      & 10  & 42.4\% & 813K & 18.7\% & 1.6M \\
    \midrule
    \multirow{3}{*}{Bottom \%}
      & 3  & 40.6\% & 254K & 23.0\% & 470K \\
      & 5  & 42.6\% & 409K & 19.7\% & 842K \\
      & 10  & 40.6\% & 835K & 18.3\% & 1.7M \\
    \bottomrule
  \end{tabular}
\end{table}

\section{GoBrowse Results}
\label{app:gobrowse_results}

We additionally evaluate on 341 tasks sampled from GoBrowse \citep{gobrowse2025}, using an LLM-as-a-judge evaluation with \JUDGEMODEL{}. GoBrowse tasks tend to be shorter (4--6 steps on average) and easier (86--90\% baselines), providing a complementary evaluation to the harder WebArena-Lite and Online-Mind2Web benchmarks. All results are averaged across 3 runs.

\subsection{Static Scaling with Majority Vote}

\tref{tab:gb_static} shows that GoBrowse exhibits similar diminishing returns to WebArena-Lite: scaling from $N{=}1$ to $N{=}10$ provides modest gains, while $N{=}20$ shows no further improvement.

\begin{table}[h]
  \centering
  \caption{\textbf{Static scaling on GoBrowse.} Majority voting success rates under varying $N$.}
  \label{tab:gb_static}
  \small
  \begin{tabular}{@{}lcc@{}}
    \toprule
    Budget & Success & Tokens \\
    \midrule
    $N{=}1$ & 86.9\% & 47K \\
    $N{=}5$ & 87.8\% & 249K \\
    $N{=}10$ & \textbf{88.0\%} & 481K \\
    $N{=}20$ & 87.8\% & 995K \\
    \bottomrule
  \end{tabular}
\end{table}

\subsection{Arbiter Scaling}

\tref{tab:gb_arbiter} shows arbiter scaling results on GoBrowse at $N{=}5$. Unlike WebArena-Lite where $K{=}20$ degrades, GoBrowse shows steady improvement with increasing $K$.

\begin{table}[h]
  \centering
  \caption{\textbf{Arbiter scaling on GoBrowse ($N{=}5$).}}
  \label{tab:gb_arbiter}
  \small
  \begin{tabular}{@{}lcc@{}}
    \toprule
    Method & Success & Tokens \\
    \midrule
    Majority vote ($N{=}5$) & 87.8\% & 249K \\
    Arbiter ($K{=}1$) & 88.6\% & 227K \\
    Arbiter scaling ($K{=}5$) & 88.2\% & 351K \\
    Arbiter scaling ($K{=}10$) & 88.7\% & 541K \\
    Arbiter scaling ($K{=}20$) & \textbf{89.6\%} & 733K \\
    \bottomrule
  \end{tabular}
\end{table}

\subsection{DeepConf on GoBrowse}

\tref{tab:gb_deepconf} presents DeepConf results. Average trace at $N{=}20$ achieves 90.3\%, the strongest DeepConf result across all three benchmarks, suggesting that token-level confidence signals are more effective on shorter, easier trajectories.

\begin{table}[h]
  \centering
  \caption{\textbf{DeepConf variants on GoBrowse.}}
  \label{tab:gb_deepconf}
  \small
  \begin{tabular}{@{}llcc@{}}
    \toprule
    Variant & $N$ & Success & Tokens \\
    \midrule
    \multirow{4}{*}{Avg.\ Trace}
      & 3  & 87.7\% & 134K \\
      & 5  & 89.4\% & 235K \\
      & 10  & 89.1\% & 475K \\
      & 20  & \textbf{90.3\%} & 968K \\
    \midrule
    \multirow{4}{*}{Tail}
      & 3  & 86.8\% & 138K \\
      & 5  & 88.3\% & 230K \\
      & 10  & 86.6\% & 452K \\
      & 20  & 88.1\% & 951K \\
    \midrule
    \multirow{4}{*}{Bottom \%}
      & 3  & 88.4\% & 145K \\
      & 5  & 88.1\% & 240K \\
      & 10  & 89.0\% & 463K \\
      & 20  & 88.2\% & 973K \\
    \bottomrule
  \end{tabular}
\end{table}

\subsection{\OURS on GoBrowse}

\tref{tab:gb_catts} presents \OURS results at $N{=}10$. Both entropy-gated and margin-gated variants improve over majority voting (88.0\%) while using fewer tokens. Margin-gated \OURS achieves 90.4\% with only 372K tokens (23\% fewer than majority voting at 481K).

\begin{table}[h]
  \centering
  \caption{\textbf{\OURS results on GoBrowse ($N{=}10$).}}
  \label{tab:gb_catts}
  \small
  \begin{tabular}{@{}lcc@{}}
    \toprule
    Method & Success & Tokens \\
    \midrule
    Majority vote & 88.0\% & 481K \\
    Always-arbitrate & 88.3\% & 443K \\
    \rowcolor{green!12}
    \OURS ($H$, $\tau{=}0.5$) & 90.2\% & 422K \\
    \rowcolor{green!12}
    \OURS ($\Delta$, $\tau{=}0.5$) & \textbf{90.4\%} & 372K \\
    \bottomrule
  \end{tabular}
\end{table}

\subsection{GoBrowse Threshold Sweep}

\tref{tab:gb_threshold_sweep} presents the full threshold sweep for \OURS on GoBrowse. Performance is robust across thresholds, with most configurations outperforming the 88.0\% majority voting baseline.

\begin{table}[h]
  \centering
  \caption{\textbf{\OURS threshold sweep on GoBrowse.} Success rates (\%) for entropy-gated (left) and margin-gated (right) at $N{=}10$.}
  \small
  \begin{subtable}[t]{0.48\linewidth}
    \centering
    \caption{Entropy-gated}
    \begin{tabular}{@{}ccccc@{}}
      \toprule
      $\tau$ & $N{=}3$ & $N{=}5$ & $N{=}10$ & $N{=}20$ \\
      \midrule
      0.2 & 88.6 & 88.9 & 90.3 & 89.4 \\
      0.3 & 88.9 & \textbf{90.3} & 90.3 & 88.9 \\
      0.4 & 88.0 & 89.7 & 89.7 & 89.7 \\
      0.5 & 88.3 & 89.7 & \textbf{91.2} & 86.8 \\
      0.6 & \textbf{88.7} & 89.3 & 87.7 & 88.9 \\
      0.7 & 88.6 & 87.7 & 90.0 & 88.0 \\
      0.8 & 88.9 & 89.4 & 89.1 & \textbf{90.9} \\
      \bottomrule
    \end{tabular}
  \end{subtable}
  \hfill
  \begin{subtable}[t]{0.48\linewidth}
    \centering
    \caption{Margin-gated}
    \begin{tabular}{@{}ccccc@{}}
      \toprule
      $\tau$ & $N{=}3$ & $N{=}5$ & $N{=}10$ & $N{=}20$ \\
      \midrule
      0.2 & 90.0 & 88.3 & 89.7 & \textbf{90.3} \\
      0.3 & 88.3 & 88.6 & 89.1 & \textbf{90.6} \\
      0.4 & 88.6 & 89.4 & 88.3 & 89.6 \\
      0.5 & 88.3 & \textbf{90.0} & \textbf{90.6} & 88.3 \\
      0.6 & \textbf{91.2} & 88.3 & 89.7 & 86.8 \\
      0.7 & 89.7 & 88.9 & 88.9 & 88.0 \\
      0.8 & 89.4 & 89.9 & 89.7 & \textbf{90.6} \\
      \bottomrule
    \end{tabular}
  \end{subtable}
  \label{tab:gb_threshold_sweep}
\end{table}

\clearpage
\section{Evaluation on Additional Model Families}
\label{app:additional_models}

This section provides additional evaluations of \OURS across different foundation model families to demonstrate the generalizability of our test-time scaling approach.

\subsection{Gemini 3.1 Flash Lite}
\label{app:gemini_evaluation}

We also conducted additional experiments to show the stability of our method across model families by evaluating several high scoring methods    on the Gemini 3.1 Flash Lite model. We ran these experiments on WebArena-Lite to directly compare scaling behavior across foundation models, with results averaged across three runs. Gemini 3.1 Flash Lite was used with the high reasoning setting.

\begin{table}[h]
  \centering
  \caption{\textbf{Gemini 3.1 Flash Lite Results Summary (WebArena-Lite).} Performance and token consumption across different test-time scaling methods. \OURS maintains its efficiency and performance advantages on the Gemini model family.}
  \label{tab:gemini_flash_results}
  \vspace{1mm}
  \small
  \begin{tabular}{@{}llcc@{}}
    \toprule
    $N$ & Method & Success Rate & Avg. Tokens Per Task \\
    \midrule
    5 & Majority Vote & 42.2\% & 420K \\
    5 & Arbiter ($K{=}1$) & 43.2\% & 415K \\
    \rowcolor{green!12}
    5 & \OURS ($\Delta$, $\tau{=}0.5$) & 44.0\% & 410K \\
    \rowcolor{green!12}
    5 & \OURS ($\Delta$, $\tau{=}0.8$) & 43.2\% & 412K \\
    \bottomrule
  \end{tabular}
\end{table}

Key observations on Gemini 3.1 Flash Lite include:
\begin{itemize}
    \item The core advantages of \OURS remain consistent when shifting from gpt-oss to Gemini.
    \item At $N{=}5$, \OURS ($\Delta$, $\tau{=}0.5$) achieves a 44.0\% success rate, outperforming both standard Majority Vote (42.2\%) and the Arbiter baseline (43.2\%), all while consuming fewer average tokens per task (410K vs.\ 420K and 415K).
\end{itemize}

\subsection{Qwen3.5-122B-A10B}
\label{app:qwen_evaluation}

To further demonstrate the generalizability of \OURS, we evaluated on Qwen3.5-122B-A10B~\citep{qwen3.5} served via vLLM with native tool calling and reasoning support. We ran these experiments on WebArena-Lite using the same evaluation protocol.

\begin{table}[h]
  \centering
  \caption{\textbf{Qwen3.5-122B-A10B Results Summary (WebArena-Lite).} Performance across different test-time scaling methods. \OURS ($\Delta$) improves over both Majority Vote and Arbiter baselines.}
  \label{tab:qwen_results}
  \vspace{1mm}
  \small
  \begin{tabular}{@{}llcc@{}}
    \toprule
    $N$ & Method & Success Rate & Avg.\ Tokens Per Task \\
    \midrule
    1 & Pass@1 & 34.5\% & 106K \\
    \midrule
    3 & Majority Vote & 33.9\% & 316K \\
    3 & Arbiter ($K{=}1$) & 36.4\% & 385K \\
    \midrule
    5 & Majority Vote & 34.5\% & 525K \\
    5 & Arbiter ($K{=}1$) & 35.8\% & 569K \\
    5 & \OURS ($\Delta$, $\tau{=}0.2$) & 38.8\% & 483K \\
    \rowcolor{green!12}
    5 & \OURS ($\Delta$, $\tau{=}0.5$) & \textbf{40.6\%} & 544K \\
    \rowcolor{green!12}
    5 & \OURS ($\Delta$, $\tau{=}0.7$) & \textbf{40.6\%} & 537K \\
    \bottomrule
  \end{tabular}
\end{table}

Key observations on Qwen3.5-122B-A10B include:
\begin{itemize}
    \item \OURS demonstrates consistent improvements on WebArena-Lite, confirming that the method is not specific to gpt-oss.
    \item At $N{=}5$, \OURS ($\Delta$, $\tau{=}0.5$) achieves a 40.6\% success rate, a +6.1 percentage point improvement over Majority Vote (34.5\%) and +4.8 over the Arbiter baseline (35.8\%).
    \item Standard Majority Vote shows no scaling benefit on this model (34.5\% at both $N{=}1$ and $N{=}5$), while \OURS successfully extracts gains from additional compute, highlighting the value of confusion-aware selection.
\end{itemize}

\section{LLM Usage Disclosure}
We used AI tools to assist with grammar, plotting code, and LaTeX editing. All research decisions and analysis are our own.

%% file: colm2026_conference.bib
@article{wang2022self,
  title={Self-consistency improves chain of thought reasoning in language models},
  author={Wang, Xuezhi and Wei, Jason and Schuurmans, Dale and Le, Quoc and Chi, Ed and Narang, Sharan and Chowdhery, Aakanksha and Zhou, Denny},
  journal={arXiv preprint arXiv:2203.11171},
  year={2022}
}

@article{naik2023diversity,
  title={Diversity of thought improves reasoning abilities of llms},
  author={Naik, Ranjita and Chandrasekaran, Varun and Yuksekgonul, Mert and Palangi, Hamid and Nushi, Besmira},
  journal={arXiv preprint arXiv:2310.07088},
  year={2023}
}

@inproceedings{wang2025ranked,
  title={Ranked voting based self-consistency of large language models},
  author={Wang, Weiqin and Wang, Yile and Huang, Hui},
  booktitle={Findings of the Association for Computational Linguistics: ACL 2025},
  pages={14410--14426},
  year={2025}
}

@article{byerly2026self,
  title={Self-Consistency Falls Short! The Adverse Effects of Positional Bias on Long-Context Problems},
  author={Byerly, Adam and Khashabi, Daniel},
  journal={Transactions of the Association for Computational Linguistics},
  volume={14},
  pages={292--317},
  year={2026},
  publisher={MIT Press}
}

@inproceedings{fu2026deep,
  title={Deep think with confidence},
  author={Fu, Yichao and Wang, Xuewei and Zhang, Hao and Tian, Yuandong and Zhao, Jiawei},
  booktitle={International Conference on Learning Representations},
  volume={2026},
  pages={94355--94377},
  year={2026}
}

@article{yao2023tree,
  title={Tree of thoughts: Deliberate problem solving with large language models},
  author={Yao, Shunyu and Yu, Dian and Zhao, Jeffrey and Shafran, Izhak and Griffiths, Tom and Cao, Yuan and Narasimhan, Karthik},
  journal={Advances in neural information processing systems},
  volume={36},
  pages={11809--11822},
  year={2023}
}

@inproceedings{yao2023react,
title={ReAct: Synergizing Reasoning and Acting in Language Models},
author={Shunyu Yao and Jeffrey Zhao and Dian Yu and Nan Du and Izhak Shafran and Karthik R Narasimhan and Yuan Cao},
booktitle={The Eleventh International Conference on Learning Representations },
year={2023},
url={https://openreview.net/forum?id=WE_vluYUL-X}
}

@inproceedings{erdogan2025plan,
  title={Plan-and-Act: Improving Planning of Agents for Long-Horizon Tasks},
  author={Erdogan, Lutfi Eren and Lee, Nicholas and Kim, Sehoon and Moon, Suhong and Furuta, Hiroki and Anumanchipalli, Gopala and Keutzer, Kurt and Gholami, Amir},
  booktitle={Forty-second International Conference on Machine Learning},
  year={2025}
}

@inproceedings{kim2024llm,
  title={An llm compiler for parallel function calling},
  author={Kim, Sehoon and Moon, Suhong and Tabrizi, Ryan and Lee, Nicholas and Mahoney, Michael W and Keutzer, Kurt and Gholami, Amir},
  booktitle={Forty-first International Conference on Machine Learning},
  year={2024}
}

@inproceedings{zhou2024webarena,
  title={Webarena: A realistic web environment for building autonomous agents},
  author = {Zhou, Shuyan and Xu, Frank F and Zhu, Hao and Zhou, Xuhui and Lo, Robert and Sridhar, Abishek and Cheng, Xianyi and Ou, Tianyue and Bisk, Yonatan and Fried, Daniel and Alon, Uri and Neubig, Graham},
  booktitle={International Conference on Learning Representations},
  volume={2024},
  pages={15585--15606},
  year={2024}
}

@article{shinn2023reflexion,
  title={Reflexion: Language agents with verbal reinforcement learning},
  author={Shinn, Noah and Cassano, Federico and Gopinath, Ashwin and Narasimhan, Karthik and Yao, Shunyu},
  journal={Advances in Neural Information Processing Systems},
  volume={36},
  pages={8634--8652},
  year={2023}
}

@misc{mdap2025,
  title         = {Solving a Million-Step LLM Task with Zero Errors},
  author        = {Meyerson, Elliot and Paolo, Giuseppe and Dailey, Roberto and Shahrzad, Hormoz and Francon, Olivier and Hayes, Conor F. and Qiu, Xin and Hodjat, Babak and Miikkulainen, Risto},
  year          = {2025},
  eprint        = {2511.09030},
  archivePrefix = {arXiv},
  url           = {https://arxiv.org/abs/2511.09030}
}

@misc{act2016,
  title         = {Adaptive Computation Time for Recurrent Neural Networks},
  author        = {Graves, Alex},
  year          = {2016},
  eprint        = {1603.08983},
  archivePrefix = {arXiv},
  url           = {https://arxiv.org/abs/1603.08983}
}

@misc{resource_rational2024,
  title         = {Reasoning Aware Self-Consistency: Leveraging Reasoning Paths for Efficient LLM Sampling},
  author        = {Wan, Guangya and Wu, Yuqi and Chen, Jie and Li, Sheng},
  year          = {2024},
  eprint        = {2408.17017},
  archivePrefix = {arXiv},
  url           = {https://arxiv.org/abs/2408.17017}
}

@inproceedings{agentbench2023_liu,
  author       = {Xiao Liu and
                  Hao Yu and
                  Hanchen Zhang and
                  Yifan Xu and
                  Xuanyu Lei and
                  Hanyu Lai and
                  Yu Gu and
                  Hangliang Ding and
                  Kaiwen Men and
                  Kejuan Yang and
                  Shudan Zhang and
                  Xiang Deng and
                  Aohan Zeng and
                  Zhengxiao Du and
                  Chenhui Zhang and
                  Sheng Shen and
                  Tianjun Zhang and
                  Yu Su and
                  Huan Sun and
                  Minlie Huang and
                  Yuxiao Dong and
                  Jie Tang},
  title        = {AgentBench: Evaluating LLMs as Agents},
  booktitle    = {The Twelfth International Conference on Learning Representations,
                  {ICLR} 2024, Vienna, Austria, May 7-11, 2024},
  publisher    = {OpenReview.net},
  year         = {2024},
  url          = {https://openreview.net/forum?id=zAdUB0aCTQ},
  bibsource    = {dblp computer science bibliography, https://dblp.org}
}

@article{rawles2023androidinthewild,
  title={Androidinthewild: A large-scale dataset for android device control},
  author={Rawles, Christopher and Li, Alice and Rodriguez, Daniel and Riva, Oriana and Lillicrap, Timothy},
  journal={Advances in Neural Information Processing Systems},
  volume={36},
  pages={59708--59728},
  year={2023}
}

@article{rawles2024androidworld,
  title={Androidworld: A dynamic benchmarking environment for autonomous agents},
  author={Rawles, Christopher and Clinckemaillie, Sarah and Chang, Yifan and Waltz, Jonathan and Lau, Gabrielle and Fair, Marybeth and Li, Alice and Bishop, William and Li, Wei and Campbell-Ajala, Folawiyo and others},
  journal={arXiv preprint arXiv:2405.14573},
  year={2024}
}

@inproceedings{li2023api,
  title={Api-bank: A comprehensive benchmark for tool-augmented llms},
  author={Li, Minghao and Zhao, Yingxiu and Yu, Bowen and Song, Feifan and Li, Hangyu and Yu, Haiyang and Li, Zhoujun and Huang, Fei and Li, Yongbin},
  booktitle={Proceedings of the 2023 conference on empirical methods in natural language processing},
  pages={3102--3116},
  year={2023}
}

@inproceedings{appagent2024_zhang,
  author       = {Chi Zhang and
                  Zhao Yang and
                  Jiaxuan Liu and
                  Yanda Li and
                  Yucheng Han and
                  Xin Chen and
                  Zebiao Huang and
                  Bin Fu and
                  Gang Yu},
  editor       = {Naomi Yamashita and
                  Vanessa Evers and
                  Koji Yatani and
                  Sharon Xianghua Ding and
                  Bongshin Lee and
                  Marshini Chetty and
                  Phoebe O. Toups Dugas},
  title        = {AppAgent: Multimodal Agents as Smartphone Users},
  booktitle    = {Proceedings of the 2025 {CHI} Conference on Human Factors in Computing
                  Systems, {CHI} 2025, YokohamaJapan, 26 April 2025- 1 May 2025},
  pages        = {70:1--70:20},
  publisher    = {{ACM}},
  year         = {2025},
  url          = {https://doi.org/10.1145/3706598.3713600},
  doi          = {10.1145/3706598.3713600},
  bibsource    = {dblp computer science bibliography, https://dblp.org}
}

@article{wei2022chain,
  title={Chain-of-thought prompting elicits reasoning in large language models},
  author = {Wei, Jason and Wang, Xuezhi and Schuurmans, Dale and Bosma, Maarten and Ichter, Brian and Xia, Fei and Chi, Ed and Le, Quoc V and Zhou, Denny},
  journal={Advances in neural information processing systems},
  volume={35},
  pages={24824--24837},
  year={2022}
}

@inproceedings{generative_agents2023_park,
  author       = {Joon Sung Park and
                  Joseph C. O'Brien and
                  Carrie Jun Cai and
                  Meredith Ringel Morris and
                  Percy Liang and
                  Michael S. Bernstein},
  editor       = {Sean Follmer and
                  Jeff Han and
                  J{\"{u}}rgen Steimle and
                  Nathalie Henry Riche},
  title        = {Generative Agents: Interactive Simulacra of Human Behavior},
  booktitle    = {Proceedings of the 36th Annual {ACM} Symposium on User Interface Software
                  and Technology, {UIST} 2023, San Francisco, CA, USA, 29 October 2023-
                  1 November 2023},
  pages        = {2:1--2:22},
  publisher    = {{ACM}},
  year         = {2023},
  url          = {https://doi.org/10.1145/3586183.3606763},
  doi          = {10.1145/3586183.3606763},
  bibsource    = {dblp computer science bibliography, https://dblp.org}
}

@article{patil2024gorilla,
  title={Gorilla: Large language model connected with massive apis},
  author={Patil, Shishir G and Zhang, Tianjun and Wang, Xin and Gonzalez, Joseph E},
  journal={Advances in Neural Information Processing Systems},
  volume={37},
  pages={126544--126565},
  year={2024}
}

@inproceedings{besta2024graph,
  title={Graph of thoughts: Solving elaborate problems with large language models},
  author={Besta, Maciej and Blach, Nils and Kubicek, Ales and Gerstenberger, Robert and Podstawski, Michal and Gianinazzi, Lukas and Gajda, Joanna and Lehmann, Tomasz and Niewiadomski, Hubert and Nyczyk, Piotr and others},
  booktitle={Proceedings of the AAAI conference on artificial intelligence},
  volume={38},
  number={16},
  pages={17682--17690},
  year={2024}
}

@article{know_what_they_know2022_kadavath,
  author       = {Saurav Kadavath and
                  Tom Conerly and
                  Amanda Askell and
                  Tom Henighan and
                  Dawn Drain and
                  Ethan Perez and
                  Nicholas Schiefer and
                  Zac Hatfield{-}Dodds and
                  Nova DasSarma and
                  Eli Tran{-}Johnson and
                  Scott Johnston and
                  Sheer El Showk and
                  Andy Jones and
                  Nelson Elhage and
                  Tristan Hume and
                  Anna Chen and
                  Yuntao Bai and
                  Sam Bowman and
                  Stanislav Fort and
                  Deep Ganguli and
                  Danny Hernandez and
                  Josh Jacobson and
                  Jackson Kernion and
                  Shauna Kravec and
                  Liane Lovitt and
                  Kamal Ndousse and
                  Catherine Olsson and
                  Sam Ringer and
                  Dario Amodei and
                  Tom Brown and
                  Jack Clark and
                  Nicholas Joseph and
                  Ben Mann and
                  Sam McCandlish and
                  Chris Olah and
                  Jared Kaplan},
  title        = {Language Models (Mostly) Know What They Know},
  journal      = {CoRR},
  volume       = {abs/2207.05221},
  year         = {2022},
  url          = {https://doi.org/10.48550/arXiv.2207.05221},
  doi          = {10.48550/ARXIV.2207.05221},
  eprinttype    = {arXiv},
  eprint       = {2207.05221},
  bibsource    = {dblp computer science bibliography, https://dblp.org}
}

@inproceedings{lats2024_zhou,
  author       = {Andy Zhou and
                  Kai Yan and
                  Michal Shlapentokh{-}Rothman and
                  Haohan Wang and
                  Yu{-}Xiong Wang},
  title        = {Language Agent Tree Search Unifies Reasoning, Acting, and Planning
                  in Language Models},
  booktitle    = {Forty-first International Conference on Machine Learning, {ICML} 2024,
                  Vienna, Austria, July 21-27, 2024},
  publisher    = {OpenReview.net},
  year         = {2024},
  url          = {https://openreview.net/forum?id=njwv9BsGHF},
  bibsource    = {dblp computer science bibliography, https://dblp.org}
}

@article{deng2023mind2web,
  title={Mind2web: Towards a generalist agent for the web},
  author={Deng, Xiang and Gu, Yu and Zheng, Boyuan and Chen, Shijie and Stevens, Sam and Wang, Boshi and Sun, Huan and Su, Yu},
  journal={Advances in Neural Information Processing Systems},
  volume={36},
  pages={28091--28114},
  year={2023}
}

@article{mrkl2022_karpas,
  author       = {Ehud Karpas and
                  Omri Abend and
                  Yonatan Belinkov and
                  Barak Lenz and
                  Opher Lieber and
                  Nir Ratner and
                  Yoav Shoham and
                  Hofit Bata and
                  Yoav Levine and
                  Kevin Leyton{-}Brown and
                  Dor Muhlgay and
                  Noam Rozen and
                  Erez Schwartz and
                  Gal Shachaf and
                  Shai Shalev{-}Shwartz and
                  Amnon Shashua and
                  Moshe Tennenholtz},
  title        = {{MRKL} Systems: {A} modular, neuro-symbolic architecture that combines
                  large language models, external knowledge sources and discrete reasoning},
  journal      = {CoRR},
  volume       = {abs/2205.00445},
  year         = {2022},
  url          = {https://doi.org/10.48550/arXiv.2205.00445},
  doi          = {10.48550/ARXIV.2205.00445},
  eprinttype    = {arXiv},
  eprint       = {2205.00445},
  bibsource    = {dblp computer science bibliography, https://dblp.org}
}

@inproceedings{multiagent_debate2023_du,
  author       = {Yilun Du and
                  Shuang Li and
                  Antonio Torralba and
                  Joshua B. Tenenbaum and
                  Igor Mordatch},
  title        = {Improving Factuality and Reasoning in Language Models through Multiagent
                  Debate},
  booktitle    = {Forty-first International Conference on Machine Learning, {ICML} 2024,
                  Vienna, Austria, July 21-27, 2024},
  publisher    = {OpenReview.net},
  year         = {2024},
  url          = {https://openreview.net/forum?id=zj7YuTE4t8},
  bibsource    = {dblp computer science bibliography, https://dblp.org}
}

@article{xie2024osworld,
  title={Osworld: Benchmarking multimodal agents for open-ended tasks in real computer environments},
  author={Xie, Tianbao and Zhang, Danyang and Chen, Jixuan and Li, Xiaochuan and Zhao, Siheng and Cao, Ruisheng and Hua, Toh J and Cheng, Zhoujun and Shin, Dongchan and Lei, Fangyu and others},
  journal={Advances in Neural Information Processing Systems},
  volume={37},
  pages={52040--52094},
  year={2024}
}

@inproceedings{pal2022_gao,
  author       = {Luyu Gao and
                  Aman Madaan and
                  Shuyan Zhou and
                  Uri Alon and
                  Pengfei Liu and
                  Yiming Yang and
                  Jamie Callan and
                  Graham Neubig},
  editor       = {Andreas Krause and
                  Emma Brunskill and
                  Kyunghyun Cho and
                  Barbara Engelhardt and
                  Sivan Sabato and
                  Jonathan Scarlett},
  title        = {{PAL:} Program-aided Language Models},
  booktitle    = {International Conference on Machine Learning, {ICML} 2023, 23-29 July
                  2023, Honolulu, Hawaii, {USA}},
  series       = {Proceedings of Machine Learning Research},
  volume       = {202},
  pages        = {10764--10799},
  publisher    = {{PMLR}},
  year         = {2023},
  url          = {https://proceedings.mlr.press/v202/gao23f.html},
  bibsource    = {dblp computer science bibliography, https://dblp.org}
}

@article{program_of_thoughts2022_chen,
  author       = {Wenhu Chen and
                  Xueguang Ma and
                  Xinyi Wang and
                  William W. Cohen},
  title        = {Program of Thoughts Prompting: Disentangling Computation from Reasoning
                  for Numerical Reasoning Tasks},
  journal      = {Trans. Mach. Learn. Res.},
  volume       = {2023},
  year         = {2023},
  url          = {https://openreview.net/forum?id=YfZ4ZPt8zd},
  bibsource    = {dblp computer science bibliography, https://dblp.org}
}

@inproceedings{rarr2023_gao,
  author       = {Luyu Gao and
                  Zhuyun Dai and
                  Panupong Pasupat and
                  Anthony Chen and
                  Arun Tejasvi Chaganty and
                  Yicheng Fan and
                  Vincent Y. Zhao and
                  Ni Lao and
                  Hongrae Lee and
                  Da{-}Cheng Juan and
                  Kelvin Guu},
  editor       = {Anna Rogers and
                  Jordan L. Boyd{-}Graber and
                  Naoaki Okazaki},
  title        = {{RARR:} Researching and Revising What Language Models Say, Using Language
                  Models},
  booktitle    = {Proceedings of the 61st Annual Meeting of the Association for Computational
                  Linguistics (Volume 1: Long Papers), {ACL} 2023, Toronto, Canada,
                  July 9-14, 2023},
  pages        = {16477--16508},
  publisher    = {Association for Computational Linguistics},
  year         = {2023},
  url          = {https://doi.org/10.18653/v1/2023.acl-long.910},
  doi          = {10.18653/V1/2023.ACL-LONG.910},
  bibsource    = {dblp computer science bibliography, https://dblp.org}
}

@article{rewoo2023_zhou,
  author       = {Binfeng Xu and
                  Zhiyuan Peng and
                  Bowen Lei and
                  Subhabrata Mukherjee and
                  Yuchen Liu and
                  Dongkuan Xu},
  title        = {ReWOO: Decoupling Reasoning from Observations for Efficient Augmented
                  Language Models},
  journal      = {CoRR},
  volume       = {abs/2305.18323},
  year         = {2023},
  url          = {https://doi.org/10.48550/arXiv.2305.18323},
  doi          = {10.48550/ARXIV.2305.18323},
  eprinttype    = {arXiv},
  eprint       = {2305.18323},
  bibsource    = {dblp computer science bibliography, https://dblp.org}
}

@article{zelikman2022star,
  title={Star: Bootstrapping reasoning with reasoning},
  author={Zelikman, Eric and Wu, Yuhuai and Mu, Jesse and Goodman, Noah},
  journal={Advances in Neural Information Processing Systems},
  volume={35},
  pages={15476--15488},
  year={2022}
}

@inproceedings{tinyclick2024_pawlowski,
  author       = {Pawel Pawlowski and
                  Krystian Zawistowski and
                  Wojciech Lapacz and
                  Adam Wiacek and
                  Marcin Skorupa and
                  Sebastien Postansque and
                  Jakub Hoscilowicz},
  editor       = {Odette Scharenborg and
                  Catharine Oertel and
                  Khiet Truong},
  title        = {TinyClick: Single-Turn Agent for Empowering {GUI} Automation},
  booktitle    = {26th Annual Conference of the International Speech Communication Association,
                  Interspeech 2025, Rotterdam, The Netherlands, 17-21 August 2025},
  publisher    = {{ISCA}},
  year         = {2025},
  url          = {https://doi.org/10.21437/Interspeech.2025-176},
  doi          = {10.21437/INTERSPEECH.2025-176},
  bibsource    = {dblp computer science bibliography, https://dblp.org}
}

@article{schick2023toolformer,
  title={Toolformer: Language models can teach themselves to use tools},
  author={Schick, Timo and Dwivedi-Yu, Jane and Dess{\`\i}, Roberto and Raileanu, Roberta and Lomeli, Maria and Hambro, Eric and Zettlemoyer, Luke and Cancedda, Nicola and Scialom, Thomas},
  journal={Advances in Neural Information Processing Systems},
  volume={36},
  pages={68539--68551},
  year={2023}
}

@inproceedings{toolllm2023_qin,
  author       = {Yujia Qin and
                  Shihao Liang and
                  Yining Ye and
                  Kunlun Zhu and
                  Lan Yan and
                  Yaxi Lu and
                  Yankai Lin and
                  Xin Cong and
                  Xiangru Tang and
                  Bill Qian and
                  Sihan Zhao and
                  Lauren Hong and
                  Runchu Tian and
                  Ruobing Xie and
                  Jie Zhou and
                  Mark Gerstein and
                  Dahai Li and
                  Zhiyuan Liu and
                  Maosong Sun},
  title        = {ToolLLM: Facilitating Large Language Models to Master 16000+ Real-world
                  APIs},
  booktitle    = {The Twelfth International Conference on Learning Representations,
                  {ICLR} 2024, Vienna, Austria, May 7-11, 2024},
  publisher    = {OpenReview.net},
  year         = {2024},
  url          = {https://openreview.net/forum?id=dHng2O0Jjr},
  bibsource    = {dblp computer science bibliography, https://dblp.org}
}

@article{turpin2023language,
  title={Language models don't always say what they think: Unfaithful explanations in chain-of-thought prompting},
  author={Turpin, Miles and Michael, Julian and Perez, Ethan and Bowman, Samuel},
  journal={Advances in Neural Information Processing Systems},
  volume={36},
  pages={74952--74965},
  year={2023}
}

@inproceedings{visualwebarena2024_koh,
  author       = {Jing Yu Koh and
                  Robert Lo and
                  Lawrence Jang and
                  Vikram Duvvur and
                  Ming Chong Lim and
                  Po{-}Yu Huang and
                  Graham Neubig and
                  Shuyan Zhou and
                  Russ Salakhutdinov and
                  Daniel Fried},
  editor       = {Lun{-}Wei Ku and
                  Andre Martins and
                  Vivek Srikumar},
  title        = {VisualWebArena: Evaluating Multimodal Agents on Realistic Visual Web
                  Tasks},
  booktitle    = {Proceedings of the 62nd Annual Meeting of the Association for Computational
                  Linguistics (Volume 1: Long Papers), {ACL} 2024, Bangkok, Thailand,
                  August 11-16, 2024},
  pages        = {881--905},
  publisher    = {Association for Computational Linguistics},
  year         = {2024},
  url          = {https://doi.org/10.18653/v1/2024.acl-long.50},
  doi          = {10.18653/V1/2024.ACL-LONG.50},
  bibsource    = {dblp computer science bibliography, https://dblp.org}
}

@article{voyager2023_wang,
  author       = {Guanzhi Wang and
                  Yuqi Xie and
                  Yunfan Jiang and
                  Ajay Mandlekar and
                  Chaowei Xiao and
                  Yuke Zhu and
                  Linxi Fan and
                  Anima Anandkumar},
  title        = {Voyager: An Open-Ended Embodied Agent with Large Language Models},
  journal      = {Trans. Mach. Learn. Res.},
  volume       = {2024},
  year         = {2024},
  url          = {https://openreview.net/forum?id=ehfRiF0R3a},
  bibsource    = {dblp computer science bibliography, https://dblp.org}
}

@article{webgpt2021_nakano,
  author       = {Reiichiro Nakano and
                  Jacob Hilton and
                  Suchir Balaji and
                  Jeff Wu and
                  Long Ouyang and
                  Christina Kim and
                  Christopher Hesse and
                  Shantanu Jain and
                  Vineet Kosaraju and
                  William Saunders and
                  Xu Jiang and
                  Karl Cobbe and
                  Tyna Eloundou and
                  Gretchen Krueger and
                  Kevin Button and
                  Matthew Knight and
                  Benjamin Chess and
                  John Schulman},
  title        = {WebGPT: Browser-assisted question-answering with human feedback},
  journal      = {CoRR},
  volume       = {abs/2112.09332},
  year         = {2021},
  url          = {https://arxiv.org/abs/2112.09332},
  eprinttype    = {arXiv},
  eprint       = {2112.09332},
  bibsource    = {dblp computer science bibliography, https://dblp.org}
}

@article{yao2022webshop,
  title={Webshop: Towards scalable real-world web interaction with grounded language agents},
  author={Yao, Shunyu and Chen, Howard and Yang, John and Narasimhan, Karthik},
  journal={Advances in Neural Information Processing Systems},
  volume={35},
  pages={20744--20757},
  year={2022}
}

@inproceedings{workarena2024,
  author       = {Alexandre Drouin and
                  Maxime Gasse and
                  Massimo Caccia and
                  Issam H. Laradji and
                  Manuel Del Verme and
                  Tom Marty and
                  David V{\'{a}}zquez and
                  Nicolas Chapados and
                  Alexandre Lacoste},
  title        = {WorkArena: How Capable are Web Agents at Solving Common Knowledge
                  Work Tasks?},
  booktitle    = {Forty-first International Conference on Machine Learning, {ICML} 2024,
                  Vienna, Austria, July 21-27, 2024},
  publisher    = {OpenReview.net},
  year         = {2024},
  url          = {https://openreview.net/forum?id=BRfqYrikdo},
  bibsource    = {dblp computer science bibliography, https://dblp.org}
}

@inproceedings{workflow_guided_exploration2018_liu,
  author       = {Evan Zheran Liu and
                  Kelvin Guu and
                  Panupong Pasupat and
                  Tianlin Shi and
                  Percy Liang},
  title        = {Reinforcement Learning on Web Interfaces using Workflow-Guided Exploration},
  booktitle    = {6th International Conference on Learning Representations, {ICLR} 2018,
                  Vancouver, BC, Canada, April 30 - May 3, 2018, Conference Track Proceedings},
  publisher    = {OpenReview.net},
  year         = {2018},
  url          = {https://openreview.net/forum?id=ryTp3f-0-},
  bibsource    = {dblp computer science bibliography, https://dblp.org}
}

@inproceedings{worldofbits2017_shi,
  author       = {Tianlin Shi and
                  Andrej Karpathy and
                  Linxi Fan and
                  Jonathan Hernandez and
                  Percy Liang},
  editor       = {Doina Precup and
                  Yee Whye Teh},
  title        = {World of Bits: An Open-Domain Platform for Web-Based Agents},
  booktitle    = {Proceedings of the 34th International Conference on Machine Learning,
                  {ICML} 2017, Sydney, NSW, Australia, 6-11 August 2017},
  series       = {Proceedings of Machine Learning Research},
  volume       = {70},
  pages        = {3135--3144},
  publisher    = {{PMLR}},
  year         = {2017},
  url          = {http://proceedings.mlr.press/v70/shi17a.html},
  bibsource    = {dblp computer science bibliography, https://dblp.org}
}

@article{kojima2022large,
  title={Large language models are zero-shot reasoners},
  author={Kojima, Takeshi and Gu, Shixiang Shane and Reid, Machel and Matsuo, Yutaka and Iwasawa, Yusuke},
  journal={Advances in neural information processing systems},
  volume={35},
  pages={22199--22213},
  year={2022}
}

@InProceedings{bagel2024_yang,
  title = {{BAGEL}: Bootstrapping Agents by Guiding Exploration with Language},
  author = {Murty, Shikhar and Manning, Christopher D and Shaw, Peter and Joshi, Mandar and Lee, Kenton},
  booktitle = {Proceedings of the 41st International Conference on Machine Learning},
  pages = {36894--36910},
  year = {2024},
  editor = {Salakhutdinov, Ruslan and Kolter, Zico and Heller, Katherine and Weller, Adrian and Oliver, Nuria and Scarlett, Jonathan and Berkenkamp, Felix},
  volume = {235},
  series = {Proceedings of Machine Learning Research},
  month = {21--27 Jul},
  publisher = {PMLR},
  url = {https://proceedings.mlr.press/v235/murty24a.html}
}

@inproceedings{stabletoolbench2024_guo,
  title = "StableToolBench: Towards Stable Large-Scale Benchmarking on Tool Learning of Large Language Models",
  author = "Guo, Zhicheng and Cheng, Sijie and Wang, Hao and Liang, Shihao and Qin, Yujia and Li, Peng and Liu, Zhiyuan and Sun, Maosong and Liu, Yang",
  booktitle = "Findings of the Association for Computational Linguistics: ACL 2024",
  month = aug,
  year = "2024",
  address = "Bangkok, Thailand",
  publisher = "Association for Computational Linguistics",
  pages = "11143--11156",
  url = "https://aclanthology.org/2024.findings-acl.664/"
}

@misc{workarena_pp2024,
  title={WorkArena++: Towards Compositional Planning and Reasoning-based Common Knowledge Work Tasks},
  author={Boisvert, Léo and Thakkar, Megh and Gasse, Maxime and Caccia, Massimo and Le Sellier De Chezelles, Thibault and Cappart, Quentin and Chapados, Nicolas and Lacoste, Alexandre and Drouin, Alexandre},
  year={2024},
  eprint={2407.05291},
  archivePrefix={arXiv},
  primaryClass={cs.AI},
  url={https://arxiv.org/abs/2407.05291}
}

@inproceedings{erdogan2024tinyagent,
  title={Tinyagent: Function calling at the edge},
  author={Erdogan, Lutfi Eren and Lee, Nicholas and Jha, Siddharth and Kim, Sehoon and Tabrizi, Ryan and Moon, Suhong and Hooper, Coleman Richard Charles and Anumanchipalli, Gopala and Keutzer, Kurt and Gholami, Amir},
  booktitle={Proceedings of the 2024 Conference on Empirical Methods in Natural Language Processing: System Demonstrations},
  pages={80--88},
  year={2024}
}

@misc{gobrowse2025,
  title={Go-Browse: Training Web Agents with Structured Exploration},
  author={Apurva Gandhi and Graham Neubig},
  year={2025},
  eprint={2506.03533},
  archivePrefix={arXiv},
  primaryClass={cs.CL},
  url={https://arxiv.org/abs/2506.03533}
}

@misc{rsa2025,
  title={Recursive Self-Aggregation Unlocks Deep Thinking in Large Language Models},
  author={Siddarth Venkatraman and Vineet Jain and Sarthak Mittal and Vedant Shah and Johan Obando-Ceron and Yoshua Bengio and Brian R. Bartoldson and Bhavya Kailkhura and Guillaume Lajoie and Glen Berseth and Nikolay Malkin and Moksh Jain},
  year={2025},
  eprint={2509.26626},
  archivePrefix={arXiv},
  primaryClass={cs.LG},
  url={https://arxiv.org/abs/2509.26626}
}

@article{brown2024large,
  title={Large language monkeys: Scaling inference compute with repeated sampling},
  author={Brown, Bradley and Juravsky, Jordan and Ehrlich, Ryan and Clark, Ronald and Le, Quoc V and R{\'e}, Christopher and Mirhoseini, Azalia},
  journal={arXiv preprint arXiv:2407.21787},
  year={2024}
}

@article{cobbe2021training,
  title={Training verifiers to solve math word problems},
  author={Cobbe, Karl and Kosaraju, Vineet and Bavarian, Mohammad and Chen, Mark and Jun, Heewoo and Kaiser, Lukasz and Plappert, Matthias and Tworek, Jerry and Hilton, Jacob and Nakano, Reiichiro and others},
  journal={arXiv preprint arXiv:2110.14168},
  year={2021}
}

@article{snell2024scaling,
  title={Scaling llm test-time compute optimally can be more effective than scaling model parameters},
  author={Snell, Charlie and Lee, Jaehoon and Xu, Kelvin and Kumar, Aviral},
  journal={arXiv preprint arXiv:2408.03314},
  year={2024}
}

@inproceedings{muennighoff2025s1,
  title={s1: Simple test-time scaling},
  author={Muennighoff, Niklas and Yang, Zitong and Shi, Weijia and Li, Xiang Lisa and Fei-Fei, Li and Hajishirzi, Hannaneh and Zettlemoyer, Luke and Liang, Percy and Cand{\`e}s, Emmanuel and Hashimoto, Tatsunori B},
  booktitle={Proceedings of the 2025 Conference on Empirical Methods in Natural Language Processing},
  pages={20286--20332},
  year={2025}
}

@article{zhang2024generative,
  title={Generative verifiers: Reward modeling as next-token prediction},
  author={Zhang, Lunjun and Hosseini, Arian and Bansal, Hritik and Kazemi, Mehran and Kumar, Aviral and Agarwal, Rishabh},
  journal={arXiv preprint arXiv:2408.15240},
  year={2024}
}

@article{cuadron2025danger,
  title={The danger of overthinking: Examining the reasoning-action dilemma in agentic tasks},
  author={Cuadron, Alejandro and Li, Dacheng and Ma, Wenjie and Wang, Xingyao and Wang, Yichuan and Zhuang, Siyuan and Liu, Shu and Schroeder, Luis Gaspar and Xia, Tian and Mao, Huanzhi and others},
  journal={arXiv preprint arXiv:2502.08235},
  year={2025}
}

@article{zheng2023judging,
  title={Judging llm-as-a-judge with mt-bench and chatbot arena},
  author={Zheng, Lianmin and Chiang, Wei-Lin and Sheng, Ying and Zhuang, Siyuan and Wu, Zhanghao and Zhuang, Yonghao and Lin, Zi and Li, Zhuohan and Li, Dacheng and Xing, Eric and others},
  journal={Advances in neural information processing systems},
  volume={36},
  pages={46595--46623},
  year={2023}
}

@inproceedings{xue2025an,
title={An Illusion of Progress? Assessing the Current State of Web Agents},
author={Tianci Xue and Weijian Qi and Tianneng Shi and Chan Hee Song and Boyu Gou and Dawn Song and Huan Sun and Yu Su},
booktitle={Second Conference on Language Modeling},
year={2025},
url={https://openreview.net/forum?id=6jZi4HSs6o}
}

@misc{qwen3.5,
  title   = {Qwen3.5: Towards Native Multimodal Agents},
  author  = {{Qwen Team}},
  year    = {2026},
  url     = {https://qwen.ai/blog?id=qwen3.5},
}

@inproceedings{liu2025visualagentbench,
  title={Visualagentbench: Towards large multimodal models as visual foundation agents},
  author={Liu, Xiao and Zhang, Tianjie and Gu, Yu and Iong, Iat Long and XiXuan, Song and Xu, Yifan and Zhang, Shudan and Lai, Hanyu and Sun, Jiadai and Yang, Xinyue and others},
  booktitle={International Conference on Learning Representations},
  volume={2025},
  pages={95650--95707},
  year={2025}
}
